\documentclass[3p, preprint, 12pt]{elsarticle}

\usepackage{graphicx}
\usepackage{amssymb}
\usepackage{lineno}
\usepackage{threeparttable}
\usepackage{amsmath}
\usepackage{mathtools}
\usepackage{bm}
\usepackage{subcaption}
\usepackage{booktabs}
\usepackage{hyperref}
\usepackage{multirow}
\usepackage{xcolor}
\usepackage[ruled,vlined]{algorithm2e}
\usepackage{hyperref}
\usepackage{enumitem}
\usepackage{makecell}

\setlist[itemize]{noitemsep, nolistsep}
\setlist[enumerate]{noitemsep, nolistsep}

\journal{Transportation Research Part C: Emerging Technologies}

\begin{document}

\begin{frontmatter}


\title{A deep inverse reinforcement learning approach to route choice modeling with context-dependent rewards}




\author[HKUDUAP,HKUIDS]{Zhan Zhao \corref{cor1}\fnref{firstfoot}}

\author[HKUDUAP]{Yuebing Liang \fnref{firstfoot}}

\address[HKUDUAP]{
Department of Urban Planning and Design, The University of Hong Kong, Hong Kong SAR, China}
\address[HKUIDS]{
Musketeers Foundation Institute of Data Science, The University of Hong Kong, Hong Kong SAR, China}

\fntext[firstfoot]{The authors contribute equally to this paper.}
\cortext[cor1]{Corresponding author (zhanzhao@hku.hk)}

\begin{abstract}
Route choice modeling is a fundamental task in transportation planning and demand forecasting. Classical methods generally adopt the discrete choice model (DCM) framework with linear utility functions and high-level route characteristics. While several recent studies have started to explore the applicability of deep learning for route choice modeling, they are limited to path-based models with relatively simple model architectures and relying on predefined choice sets. Existing link-based models can capture the dynamic nature of link choices within the trip without the need for choice set generation, but still assume linear relationships and link-additive features. To address these issues, this study proposes a general deep inverse reinforcement learning (IRL) framework for link-based route choice modeling, which is capable of incorporating diverse features (of the state, action and trip context) and capturing complex relationships. Specifically, we adapt an adversarial IRL model to the route choice problem for efficient estimation of context-dependent reward functions without value iteration. Experiment results based on taxi GPS data from Shanghai, China validate the superior prediction performance of the proposed model over conventional DCMs and other imitation learning baselines, even for destinations unseen in the training data. Further analysis show that the model exhibits competitive computational efficiency and reasonable interpretability. The proposed methodology provides a new direction for future development of route choice models. It is general and can be adaptable to other route choice problems across different modes and networks.
\end{abstract}


\end{frontmatter}

%\linenumbers

\section{Introduction} \label{sec:intro}
To tackle traffic congestion and plan transportation infrastructure, it is important to understand, and ultimately predict, how individuals move between places through transportation networks and how various factors affect their decisions. The increasing popularity of devices equipped with location sensors offers unprecedented possibilities for studying detailed human movements and better comprehending individual routing preferences. This requires specialized analytical methods to extract human routing behavior patterns from large-scale trajectory data. Route choice modeling, as a fundamental step in transportation planning and forecasting, is the process of estimating the likely paths that individuals follow during their journeys \citep{prato_route_2009}. Empirical studies have repeatedly shown that chosen routes often deviate significantly from the shortest paths \citep{jan_using_2000,lima_understanding_2016}. Therefore, route choice models are needed to appraise perceptions of route attributes, forecast routing behavior under hypothetical scenarios, predict future traffic conditions on transportation networks, and potentially guide us to design better urban infrastructure. 

Conventional route choice models adopt the discrete choice model (DCM) framework and can be generally categorized into two types. The most common type is \textit{path-based} in the sense that the model describes a discrete choice among paths, where each path is a sequence of network links that connects an origin to a destination. Path-based models are simple but require sampling potential paths to form a finite choice set, which is a non-trivial task in a large and complex network (e.g., urban road networks). The other type is \textit{link-based} where the route choice problem is formulated as a sequence of link choices. Link-based models have several advantages. First, they do not require choice set generation and tend to produce more consistent model estimates \citep{fosgerau_link_2013}. Second, they are generally more suited to incorporate detailed link-level features (e.g., bike lanes for cyclists), which is useful for evaluating transportation network design. Third, they can be easily extended for dynamic routing situations when travelers may change route choices en route in response to updated link features (e.g., traffic congestion for drivers). Despite of these potential advantages, existing link-level models, like most DCMs, still assume simple linear utility functions and require link-additive features, making it difficult to uncover complex human routing preferences that may be nonlinear and vary by context. Therefore, we need a more flexible link-based modeling framework to incorporate context information and capture such complex routing preferences.

In recent years, deep learning has emerged as a powerful alternative to classical DCMs. Because of their multilayer structure and diverse architecture design, deep neural networks (DNNs) can capture nonlinear relationships and incorporate high-dimensional features, and have shown state-of-the-art performance in many problems, especially for mode choice modeling \citep{cantarella_multilayer_2005,wang_deep_2020}. There are relatively few studies applying DNNs for route choice modeling, and all are path-based \citep{marra_deep_2021}. Link-based models typically need to adopt a sequential structure to capture the behavioral patterns in a series of link choices. Although several recent works demonstrated the potential of sequential DNNs for trajectory prediction \citep{liang_nettraj_2022} and trajectory generation \citep{choi_trajgail_2021}, they cannot be directly applied to route choice modeling, where reaching a specific destination is necessary. As route choice models are commonly used to understand human routing preferences and make travel predictions for planning purposes, they require good interpretability and generalizability, which are often lacking in deep learning methods. This study aims to fill these research gaps.

In this study, we propose a deep inverse reinforcement learning (IRL) framework for link-based route choice modeling. IRL is well suited because it is structurally similar to dynamic DCMs \citep{rust_optimal_1987}, behaviorally interpretable, and flexible enough to incorporate deep architectures and high-dimensional features. In this framework, the link-based route choice problem is formulated as a Markov Decision Process (MDP), and the goal is to recover the underlying reward function (similar to utility functions) from observed human trajectories. Both the reward and policy functions are context-dependent and can be approximated using DNNs. Specifically, the adversarial IRL, or AIRL \citep{fu_learning_2018}, is adapted to the route choice problem to learn these functions from observed trajectories in a model-free fashion, without the need for value iteration. Several alternatives of the proposed model are also tested, including a simpler behavioral cloning (BC) method and Generative Adversarial Imitation Learning (GAIL) \citep{ho_generative_2016}. Extensive experiments are conducted using taxi GPS data on a selected road network from Shanghai, China, and the results validate the improved prediction performance of the proposed model over classical DCMs as well as other DNN baselines. The improvement holds for limited training data and unseen destinations, demonstrating the generalizability of the model. The specific contributions of this study include (1) a general deep IRL framework for route choice modeling with context-dependent reward and policy functions, (2) a specific context-dependent AIRL architecture to learn routing preferences from observed trajectories without value iteration, and (3) empirical findings for comparison across a diverse range of route choice models (path-based vs link based, linear vs nonlinear, model-based vs model-free, IRL vs imitation learning, etc.) regarding their prediction performance, computational efficiency and interpretability. While this study focuses on the driver route choice on a road network, the proposed methodology is general and should be adaptable to other route choice problems across different modes and networks. To support open science and future development of related methods, both the code and relevant data are made available online\footnote{The code and relevant data are available at https://github.com/liangchunyaobing/RCM-AIRL}.

The rest of the paper is organized as follows. Section~\ref{sec:lit} reviews the related literature on route choice models, inverse reinforcement learning, and interpretable deep learning. Section~\ref{sec:problem} presents how we formulate the route choice problem as an inverse reinforcement learning problem and how deep learning can help us solve the problem. We then introduce a specific model architecture adapted from AIRL for route choice modeling in Section~\ref{sec:rcm}. The model evaluation setup and results are described in Sections~\ref{sec:setup} and \ref{sec:results}, respectively. Finally, Section~\ref{sec:conclusion} concludes the paper by discussing the main findings, limitations, and future research directions.

\section{Literature Review} \label{sec:lit}

\subsection{Discrete Choice Models for Route Choice Modeling}
Existing route choice models generally adopt the DCM framework and fall into two types---path-based and link-based. One main challenge for path-based models is the need to account for the correlation between overlapping alternative paths, and numerous methods have been developed to address this issue, including the widely used path size logit (PSL) \citep{ben-akiva_discrete_1999}. In a real-sized network, there can be a large number of possible paths connecting each origin-destination (OD) pair. To estimate path-based models, one has to make assumptions about which paths to consider and generate a finite choice set, typically using some sort of path generation algorithms such as link elimination \citep{azevedo_algorithm_1993} and constrained enumeration \citep{prato_modeling_2007}. However, it has been shown that parameter estimates can vary significantly based on the definition of choice sets  \citep{frejinger_sampling_2009}. 

These issues with path-based models have motivated the development of link-based models in recent years. \cite{fosgerau_link_2013} proposed the recursive logit model, where a path choice is modeled as a sequence of link choices using a dynamic discrete choice framework. They found that the recursive logit is equivalent to a path-based multinomial logit model with an unrestricted choice set, but can produce more consistent parameter estimates without requiring choice set generation. Later studies proposed several extensions, including the nested recursive logit \citep{mai_nested_2015} and discounted recursive logit \citep{oyama_discounted_2017}. However, existing link-based models assume simple linear utility functions and require link-additive features \citep{zimmermann_bike_2017}, limiting their ability to uncover nonlinear effects of link/path attributes and incorporate context features about the trip/user. We need more flexible model structures to account for more complex effects of diverse features. 

\subsection{Inverse Reinforcement Learning for Route Choice Modeling}
Parallel to the development of recursive logit models, IRL has emerged as a general yet powerful framework to model sequential decision processes (e.g. routing) over the past two decades. While the goal of reinforcement learning (RL) is to learn a decision process to produce behavior that maximizes some predefined reward function, IRL inverts the problem and aims to extract a reward function from demonstration data that explains observed human behavior \citep{abbeel_apprenticeship_2004,ng_algorithms_2000}. The reward function is usually specified by a number of features, and the learnt parameters describe human preferences, which is similar to the utility function in DCMs. IRL is closely connected to imitation learning (IL), only with different learning objectives. While IRL aims to recover the reward function, IL focuses on directly learning the optimal policy from the data.

As the original IRL is prone to noise and often ambiguous, \cite{ziebart_maximum_2008} proposed the maximum entropy IRL (MaxEntIRL), which employs the principle of maximum entropy to resolve the ambiguity in choosing a distribution over decisions. In fact, it has been shown to share many similarities with the recursive logit, as both can model route choices as a sequence of link choices, assume linear reward/utility functions and can estimate a probability distribution over all feasible paths \citep{zimmermann_tutorial_2020,koch_review_2020}. One key difference is that MaxEntIRL computes all path probabilities through value iteration, while the recursive logit estimates the value functions by solving a system of linear equations \citep{fosgerau_link_2013,mai_decomposition_2018}. The former is less efficient in larger and more complex networks, and the latter is generally incompatible with nonlinear model structures.

Recent years have seen growing interest in the use of deep architectures for reward function approximation in RL/IRL \citep{mnih_human-level_2015}. \cite{wulfmeier_maximum_2016} extended the MaxEntIRL approach with DNNs, and it has been adapted for delivery route planning \citep{liu_integrating_2020} or recommendation \citep{liu_personalized_2022}, though value iteration is still required. An alternative approach, Generative Adversarial Imitation Learning (GAIL), was proposed to combine the core idea of IRL with the generative adversarial framework \cite{ho_generative_2016}. Although it has been applied to taxi driver strategy learning \citep{zhang_cgail_2020} and synthetic trajectory generation \citep{choi_trajgail_2021}, GAIL is still an IL approach, and cannot estimate specific reward functions or reveal routing preferences. More recently, a similar adversarial architecture was adopted in Adversarial Inverse Reinforcement Learning (AIRL), which is able to recover reward functions in large, high-dimensional problems \citep{fu_learning_2018}. Specifically for route choice analysis, such DNN-based IRL/IL methods have yet to be explored.

\subsection{Interpretable Deep Learning}
While it has long been recognized that DNNs can outperform DCMs for choice prediction \citep{cantarella_multilayer_2005}, a central critique of deep learning is its lack of interpretability. Understanding why a model makes a certain prediction is essential for explaining the underlying human behavior and ensure model trustworthiness. As a result, recent works have investigated theories and methods for interpretable deep learning \citep{doshi-velez_towards_2017}, though most related transportation studies focus on mode choice analysis. \cite{wang_deep_2020} demonstrated that DNNs can provide economic information as complete as classical DCMs, including utilities, elasticities and values of time. \cite{sifringer_enhancing_2020} introduced a new utility formulation by dividing the systematic utility specification into a knowledge-driven part (for interpretability) and a data-driven part (for representation learning). \cite{alwosheel_why_2021} proposed the use of activation maximization and layer-wise relevance propagation techniques to assess the validity of relationships learned in nerual networks. For mobility flow (or destination choice) analysis, \cite{simini_deep_2021} showed that DNNs can not only significantly outperform the classical gravity model, but also be interpreted using SHapley Additive exPlanations (SHAP), which estimates the contribution of each feature based on how its absence changes the model prediction \citep{lundberg_unified_2017}. This inspires us to consider a deep learning approach to route choice modeling, which can potentially improve the prediction performance while preserving most of the interpretability as in classical DCMs.

\section{Problem Formulation} \label{sec:problem}

\subsection{Route Choice as a Markov Decision Process}
The route choice problem can be regarded as a Markov decision process (MDP) \citep{ziebart_maximum_2008}, which provides a mathematical framework for modeling the sequential decision processes of an agent (or traveler). An MDP is generally defined as $M=\{S,A,T,R, \gamma\}$, where $S$ denotes the state space, $A$ the set of possible actions, $T(s,a,s')$ a transition model that determines the next state $s'\in S$ given the current state $s\in S$ and action $a\in A$, $R(s,a)$ the reward function that offers rewards to the agent based on its current state and action, and $\gamma$ the discount factor between future and present rewards. For route choice analysis, $T(s,a,s')$ is often considered to be deterministic \citep{ziebart_maximum_2008}, which means that given the current state an action will always lead to the same next state. Also, the route choice MDP is usually episodic, where the episode ends when the agent reaches the desired destination. In such cases, reward discounts are unnecessary, but we can set it slightly lower than 1 for computational efficiency. In this study, we set $\gamma$ as 0.99 following \cite{zhang_cgail_2020}. In addition, a policy $\pi (a \mid s)$ (assumed to be stochastic) gives the probability of choosing action $a$ at state $s$ under policy $\pi$. In the RL setting, the reward $R$ is given, and the objective is to find the optimal policy $\pi ^*$ that maximizes the expected cumulative reward for the agent. In the IRL setting, however, $R$ is unknown, and the objective is to recover the reward function from the observed behavior trajectories $X=\{x_1, x_2,..., x_N\}$ that are assumed to be sampled from the optimal policy $\pi ^*$. 

\begin{figure}[!ht]
  \centering
  \includegraphics[width=\textwidth]{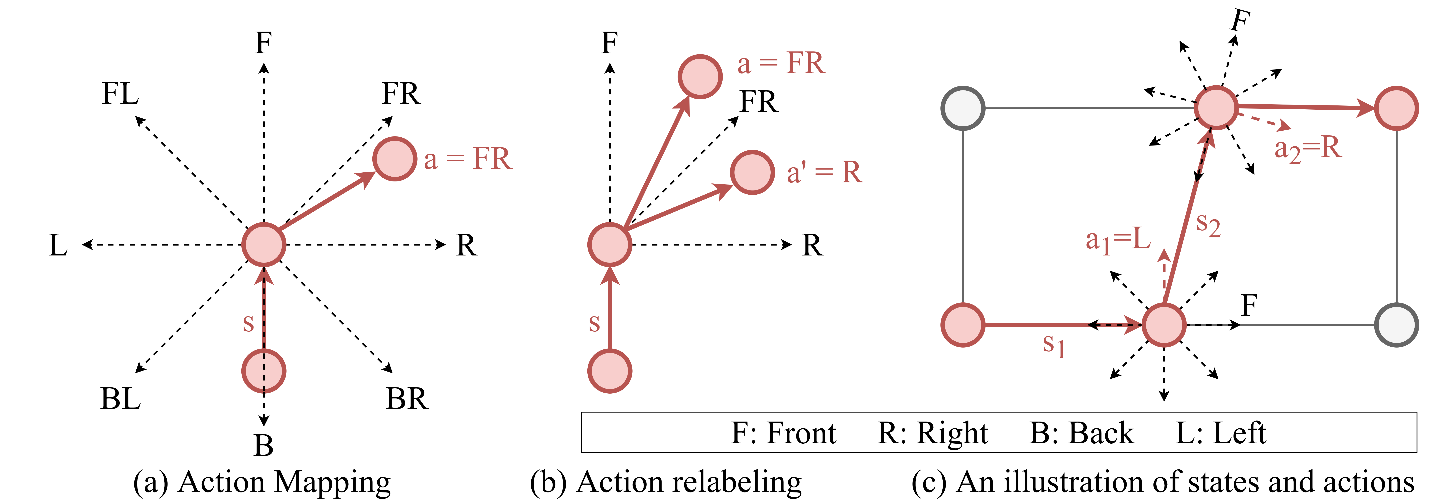}
  \caption{Definition of action space. We take two steps to define action space: action mapping (a) and action relabeling (b). First, we calculate the heading difference between each connected link pair and discretize the heading difference to 8 action (or direction) labels. If two links connecting with the same link share the same direction label, we then relabel one of them to the closest available action space.}\label{fig:action_map}
\end{figure}

In a route choice problem, the ultimate goal of an agent is to maximize their total reward (or utility), conditional on reaching the intended destination. Therefore, the trip destination significantly affects the route choice. In addition, other contextual factors that vary across trips can also have an impact, including the characteristics of the agent, trip purpose, weather, etc. To capture the influence of these factors, we define a trip-specific context vector $c$. Unlike state $s$, context $c$ is assumed to be fixed during a trip and not directly affected by traveler's actions. Specifically, we can define the state, action and context as follows: 
\begin{itemize}
    \item \textbf{State}. Each state $s\in S$ is a link in the road network indicating the current location of the traveler. Depending on data availability, the dynamic traffic condition of each link can be incorporated, but is outside the scope of this study. We will focus on the spatial aspect and use the traveler's location as the core determinant of a state, following most prior works \citep{ziebart_maximum_2008,fosgerau_link_2013}.
    \item \textbf{Action}. An action $a\in A$ indicates the link-to-link movement choice, which can be represented by either a link ID specifying the link to pass through, or a movement direction indicating the turning angle. \cite{liang_nettraj_2022} has shown that a directional representation (the latter) can yield better route prediction performance, and 8 directions are adequate to uniquely map most link-to-link movements in large road networks. Therefore, we first define a general action space $A$ consisting of 8 movement directions---forward (F), forward left (FL), left (L), backward left (BL), backward (B), backward right (BR), right (R), and forward right (FR), as shown in Figure ~\ref{fig:action_map}. Note that, while these 8 directions represent a \textit{global} set of all possible actions to take anywhere, only a subset of them are valid for most states. To reflect the local network layout, we also define a \textit{local} action space $A(s)$ to track all valid actions at each state $s$. The detailed specification of the action space is discussed in \ref{appendix:a}.
    \item \textbf{Context}. An context vector $c$ is defined to capture the metainformation of a trajectory. For route choice, this includes key trip characteristics such as the destination and purpose, as well as other contextual information including the attributes of the agent, time of day, weather, etc. These factors are likely to alter individual routing preferences in different ways, and lead to different route choice behavior. Furthermore, many of these factors are discrete and high-dimensional, with potentially nonlinear effects on routing preferences.
\end{itemize}

With the context defined, we can reasonably assume both the policy and reward functions to be context-dependent:
\begin{itemize}
    \item \textbf{Policy}. A context-dependent policy $\pi (a \mid s,c)$ determines how the agent acts at each state $s$ given context $c$. Essentially, it represents a certain routing behavior pattern. The ``optimal'' policy $\pi^*$ is basically the most representative routing pattern underlying actual human route choices. This is unknown in route choice analysis, but we can potentially infer it from observed trajectories $X$ sampled from $\pi^*$.
    \item \textbf{Reward function}. While a policy $\pi$ represents certain \textit{routing patterns}, a reward function $R(s,a \mid c)$ represents a set of \textit{routing preferences}. The latter is more fundamental, succinct, and arguably more generalizable, than the former. Similarly, the ``optimal'' reward function $R^*$ indicates the most representative routing preferences that best explain human routing behavior. This is similar to estimating the most fitting utility function in utility-based DCMs. Specifically, $R(s,a \mid c)$ can be considered as determining the utility derived from taking action $a$ at state $s$ given context $c$.
\end{itemize}

While the context can be potentially incorporated in existing link-based models with linear utility functions through the use of interaction variables, they are rarely considered in prior studies. When the number of link attributes or context dimensionality increases, the number of interaction variables needed can grow exponentially, leading to scalability issues. It is worth highlighting that, as a key element of the trip context, the destination has always been considered in existing link-based models, but the specific formulation differs from our framework. Notably, in the recursive logit \citep{fosgerau_link_2013} or MaxEntIRL \citep{ziebart_maximum_2008}, the value function of a state is also destination-dependent, as the destination acts as an absorbing state of the MDP for value function estimation. However, the instantaneous utility (or reward) of a station-action pair, $R(s, a)$, is assumed to be constant across all destinations. Our framework is different, in that we assume $R(s, a)$ to be conditional on the destination as well, which can potentially help us uncover more complex routing preferences. For example, depending on how close the agent currently is to the destination, their routing preference may be different. As an additional computational advantage, our reward formulation allows us to no longer depend on value iteration for reward estimation. It is also more flexible to account for additional contextual factors, such as the agent's characteristics, that may interact with routing preferences.

\subsection{Route Choice as an Inverse Reinforcement Learning Problem}

Recall that IRL for route choice modeling seeks to infer the reward function $R(s, a \mid c)$ given a set of observed trajectories $X=\{x_1, x_2,..., x_N\}$ that are assumed to be drawn from an optimal policy $\pi ^* (a \mid s, c)$. Note that each observed trajectory is a sequence of station-action pairs with a certain context. Let us denote the $i$-th trajectory as $x_i = \{(s^{(i)}_1, a^{(i)}_1), (s^{(i)}_2, a^{(i)}_2), ..., (s^{(i)}_{T_i}, a^{(i)}_{T_i}) \mid c^{(i)}\}$, where $(s^{(i)}_t, a^{(i)}_t)$ is the $t$-th state-action pair of $x_i$, $c^{(i)}$ the context, and $T_i$ the length of the trajectory.

In large road networks, there are many possible combinations of $(s, a, c)$, making it difficult to obtain a robust estimate of $R(s, a \mid c)$. A common solution is to approximate it with a parameterized reward function $R_\theta (s, a \mid c)$, where $\theta$ is the function parameters to be learned. The specific function approximation can take many different forms, among which DNNs are generally more flexible and adept in dealing with high-dimensional features. As the total reward of a whole trajectory $x_i$ is simply the sum of the discounted rewards for each state-action pair in the trajectory, the parameterized reward function can be expressed as
\begin{equation}
    R_{\theta}(x_i) = \sum_{t=1}^{T_i} \gamma^t R_{\theta}(s^{(i)}_t, a^{(i)}_t \mid c^{(i)}).
\end{equation}

While the underlying policy that generates the observed trajectories is assumed to be ``optimal'', actual routing behavior may be suboptimal and vary across individuals. Even for the same OD pair, different travelers may choose different routes. To address this issue, the principle of maximum entropy is adopted to handle behavioral suboptimality as well as stochasticity by operating on the distribution over possible trajectories. Following the MaxEntIRL formulation \citep{ziebart_maximum_2008}, the probability of observing any given trajectory $x$ is proportional to the exponential of its cumulative reward:
\begin{equation} \label{eq:prob}
    P_{\theta}(x) = \frac{1}{Z} \exp \big(R_{\theta}(x) \big)
\end{equation}
where the partition function $Z$ is the integral of $R_{\theta}(x)$ over all possible trajectories. Therefore, we can frame the IRL problem as solving the maximum likelihood problem based on the observed trajectories:
\begin{equation} \label{eq:mle}
    \max_{\theta} \sum_{i=1}^{N}  \log \big( P_{\theta}(x_i) \big)
\end{equation}

One main challenge to solve the above optimization problem is the computation of $Z$ in Eq.~\eqref{eq:prob}. Traditional methods are \textit{model-based}, in that they leverage the known dynamics of the routing environment to compute $Z$ exactly through value iteration \citep{ziebart_maximum_2008} or solving a system of linear equations \citep{fosgerau_link_2013,mai_decomposition_2018}. However, the former can be computationally expensive because of the large number of $(s, a, c)$ combinations in real-world routing situation, and the latter is generally incompatible with nonlinear model structures. Instead of computing $Z$ exactly, \textit{model-free} IRL methods can estimate it by learning a separate sampling distribution (or policy) that approximate the actual trajectory distribution \citep{finn_guided_2016}. Essentially, this requires us to learn both the reward and policy functions simultaneously, making it possible to recast Eq.~\eqref{eq:mle} as a generative adversarial network (GAN) optimization problem \citep{finn_connection_2016}. For specific implementation, we will focus on one of the latest such methods called Adversarial IRL (AIRL) \citep{fu_learning_2018}, and show how it can be adapted to solving the route choice problem in Section~\ref{sec:rcm}.

\section{Adversarial Inverse Reinforcement Learning for Route Choice Modeling} \label{sec:rcm}

In this section, we introduce a context-dependent AIRL architecture specifically for route choice modeling. We start by describing the original formulation of AIRL \citep{fu_learning_2018} in Section~\ref{sec:rcm:airl}, before introducing the adapted version, namely RCM-AIRL, and its specific architecture design for modeling route choice on road networks in Section~\ref{sec:rcm:main}. To show that many model design possibilities exist under the same deep learning framework, Section~\ref{sec:rcm:il} presents two additional variants of RCM-AIRL which will later be used for model comparison.

\subsection{Preliminaries: Adversarial Inverse Reinforcement Learning (AIRL)} \label{sec:rcm:airl}

In this study, we adapt an AIRL framework to inversely learn the reward and policy functions from observed trajectory data, which will allow us to recover the underlying routing preferences and predict route choice behaviors. In this section, we provide a brief introduction of AIRL, and highlight its limitations in its original form for route choice modeling, before introducing an adapted version of the model named RCM-AIRL in the next section.

AIRL is an inverse reinforcement learning algorithm based on an adversarial reward learning formulation. As in original GANs \citep{goodfellow_generative_2014}, AIRL typically consists of two models that are trained simultaneously: a generator $G$ and a discriminator $D$. The AIRL discriminator is tasked with classifying its inputs as either the output of the generator, or actual samples from the underlying data distribution. The generator, on the other hand, aims to find a policy $\pi_G$ to produce outputs that are classified by the discriminator as coming from $\pi^*$. Essentially, the generator determines the policy, and the discriminator specifies the reward function.

\cite{fu_learning_2018} showed that the trajectory-centric formulation as in \cite{finn_connection_2016} can complicate model training. Instead, it is easier to perform discrimination over state-action pairs. Therefore, the discriminator is defined as
\begin{equation}
    D_{\theta, \phi}(s,a)=\frac{\exp \big(f_{\theta, \phi}(s,a)\big)}{\exp \big(f_{\theta, \phi}(s,a)\big) + \pi_G(a \mid s)}
\end{equation}
where $f_{\theta, \phi}(s,a)$ is a function related to the reward to be learned. For a deterministic transition model as in our case, $f_{\theta, \phi}(s,a)$ is defined as
\begin{equation} \label{eq:f}
    f_{\theta, \phi}(s,a)=g_{\theta}(s,a) + \gamma h_{\phi}(s') - h_{\phi}(s)
\end{equation}
where $s'$ is the next state given the current state $s$ and action $a$, $g_{\theta}(s,a)$ is a reward approximator and $h_{\phi}(s)$ is a shaping term to mitigate the effects of unwanted reshaping on the reward approximator. At optimality, $f^*(s,a)$ is the advantage function of the optimal policy.

The objective of the discriminator $D$ is to minimize the cross-entropy loss between generated data and actual ones:
\begin{equation}
    \min_{\theta, \phi} - E_D \left[ \log \big(D_{\theta, \phi}(s, a) \big)\right] - E_{\pi_G} \left[ \log \big(1 - D_{\theta, \phi}(s, a) \big)\right]
\end{equation}
where $E_D$ and $E_{\pi_G}$ indicate expected values over actual and generated trajectories, respectively.

There are different ways to formulate the reward. In AIRL, given the discriminator $D$, \cite{fu_learning_2018} defines a modified reward function as an entropy-regularized policy objective:
\begin{equation} \label{eq:r}
    R_{\theta, \phi}(s, a) = \log \big(D_{\theta, \phi}(s, a) \big) -\log \big(1 - D_{\theta, \phi}(s, a) \big) = f_{\theta, \phi}(s,a) - \log \pi_G(a \mid s)
\end{equation}
where $f_{\theta, \phi}(s,a)$ is from Eq.~\eqref{eq:f}, and $- \log \pi_G(a \mid s)$ is added to encourage higher entropy. To be consistent with AIRL, we will use this reward function formulation in the rest of the paper. Based on this formulation, the objective of the generator $G$ is simply to find the policy $\pi_G$ that can generate trajectories with maximum reward: 
\begin{equation}
\max_{\pi_G} E_{\pi_G} \left[R_{\theta, \phi}(s, a)\right].
\end{equation}

The simultaneous training of $G$ and $D$ corresponds to a minmax two-player game, where the two models are pitted against each other \citep{goodfellow_generative_2014}. Competition in this game drives both models to improve until an optimum is reached.

The original AIRL architecture was experimented for continuous robotic control tasks and achieved state-of-the-art performance compared to several imitation learning baselines \citep{fu_learning_2018}. However, it may not work well for route choice modeling. Our preliminary analysis reveals that the trajectories generated by the standard AIRL can take a very long time to reach the desired destination, or get stuck in a loop and never reach the destination at all. The likely reason is that the reward function learned in AIRL only depends on state-action pairs (i.e., the agent's current link and movement direction), ignoring the impact of the destination. Model-based methods rely on either value iteration or solving a system of linear equations to evaluate all possible paths to the destination (as an absorbing state), and thus do not have this issue. As a model-free method, AIRL learns through trials and errors, and thus may have troubles generating sensible trajectories that reach the intended destination within a reasonable number of iterations. Fortunately, the issue can be resolved through proper model design and feature engineering. In this case, it is important to explicitly incorporate the destination information as part of the context vector $c$ in the model, which can guide the agent to make context-dependent decisions and learn more efficiently. In the next section, we will introduce a context-dependent AIRL architecture adapted for route choice modeling (RCM-AIRL).  

\subsection{RCM-AIRL: Adapting AIRL for Route Choice Modeling}\label{method:RCM-AIRL} \label{sec:rcm:main}

\begin{figure}[!ht]
  \centering
  \includegraphics[width=\textwidth]{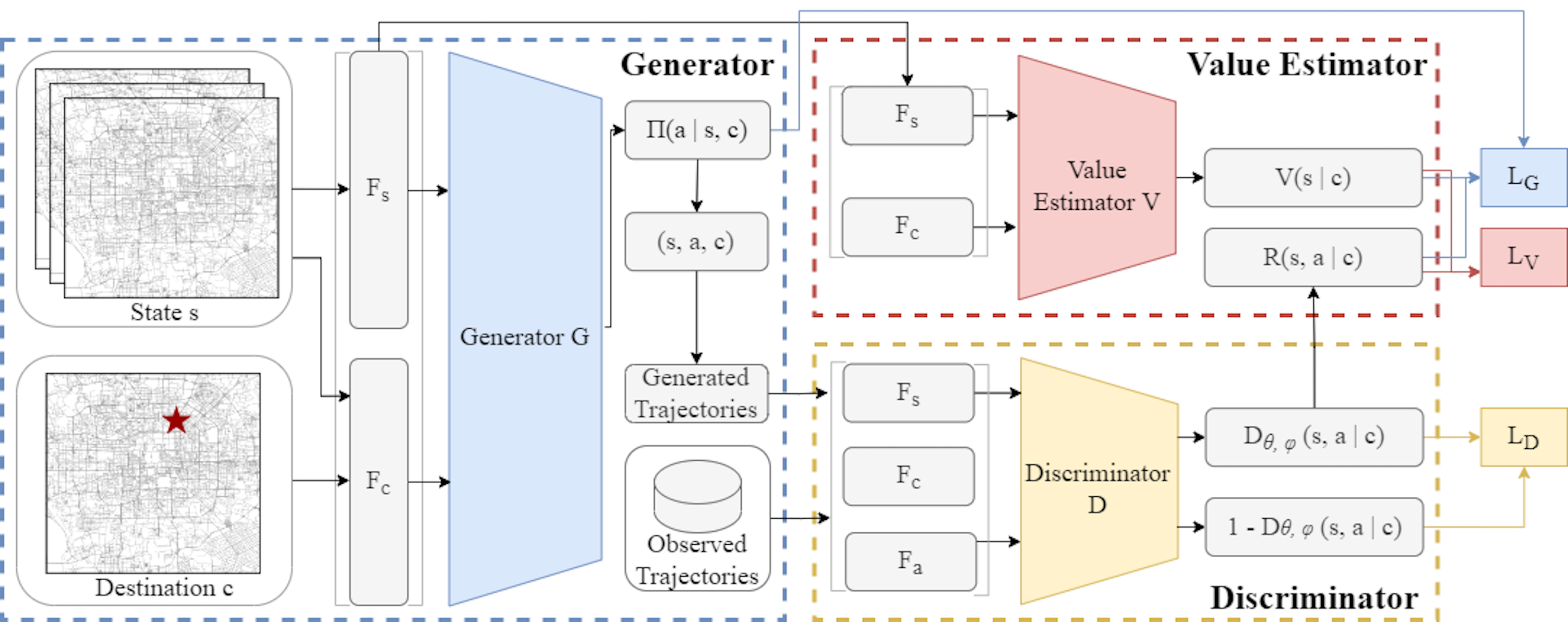}
  \caption{The model framework of RCM-AIRL}\label{fig:RCM-AIRL}
\end{figure}

The main idea behind RCM-AIRL is to have both the generator and discriminator dependent on the context features, especially those related to the destination, in addition to the features about the state and action. Figure~\ref{fig:RCM-AIRL} shows the model framework of RCM-AIRL. As in AIRL, RCM-AIRL consists of a discriminator (reward estimator) and a generator (policy estimator). We also use a value estimator to calculate the expected return of a current state to a specific destination. Below we provide more details on each module and the training algorithm of RCM-AIRL.

\subsubsection{Policy Estimator $G$ (RCM-AIRL Generator)}

\begin{figure}[!ht]
  \centering
  \includegraphics[width=\textwidth]{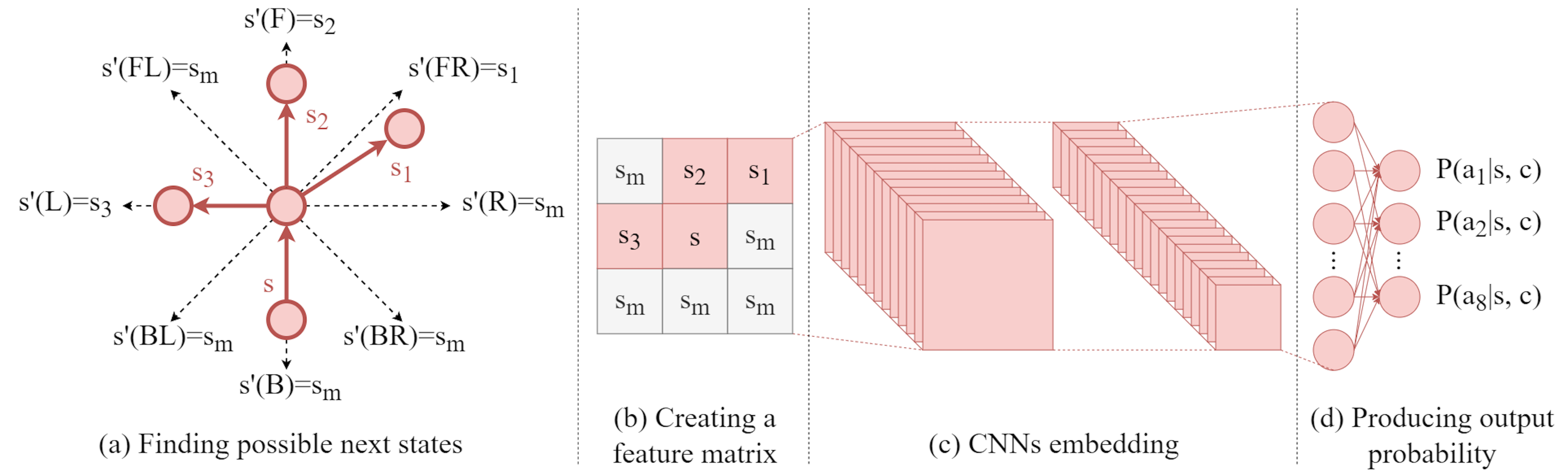}
  \caption{The network structure of the generator $G$}\label{fig:policy_network}
\end{figure}

The policy estimator $G$ aims to learn a policy $\pi_G$ that generates realistic human trajectories given an OD pair. At each step, $G$ takes two types of features as input: one is state features $F_s$, indicating features of the current link (e.g., link length); the other is context features $F_{c}$, indicating features related to the destination (e.g., the shortest path distance to the destination) and general context (e.g., the attributes of the traveler). The output of $G$ is the probability distribution for a traveler to choose different actions given the current state $s$ and context $c$. Using the origin link as the start state, the policy estimator recursively generates the next state (i.e., link) until the destination is reached. To further account for the spatial correlation between adjacent states, we also propose to use a convolutional neural network (CNN) to aggregate $F=[F_s;F_{c}]$ for both the current state and potential next states. Specifically, we denote possible next states as $S'(s)=\{s'(s, a), \forall{a} \in A(s)\}$, where $s'(s, a)$ represents the next state given the current state $s$ and action $a$. Since the road network is typically irregular, the size of $S'(s)$ can vary by the state $s$. To ensure the applicability of the policy estimator for all links in the network, we develop a network structure as illustrated in Figure~\ref{fig:policy_network}, which consists of four main steps:
\begin{enumerate}[label = (\roman*)]
    \item Find possible next states. Given a state $s$, find the next state $s'(s, a)$ via each action (i.e. direction) $a \in A$. If an action does not lead to a valid state, or $a \notin A(s)$, $s'(s, a)$ is assigned with a mask state denoted as $s_m$.
    \item Create a feature matrix. Aggregate the state and context features $F=[F_s;F_{c}]$ for $s$ and its possible next states $S'(s)$ in a $3 \times 3$ feature matrix.
    \item Learn convolutional neural networks (CNN) embedding. Use a two-layer CNN with a kernel size of 2 to learn a latent space vector from the feature matrix.
    \item Generate action probabilities. With the latent space vector learned from CNNs as input, generate outputs using a two-layer feed-forward network followed by a softmax function.
\end{enumerate}

\subsubsection{Reward Estimator $D$ (RCM-AIRL Discriminator)}
The primary goal of the discriminator is to distinguishing real human trajectories from generated ones. In RCM-AIRL, the discriminator takes the following form:
\begin{equation}
    D_{\theta, \phi}(s,a|c)=\frac{\exp \big(f_{\theta, \phi}(s,a|c)\big)}{\exp \big(f_{\theta, \phi}(s,a|c)\big) + \pi_G(a \mid s,c)}
\end{equation}
where $f_{\theta, \phi}(s,a|c)$ is defined as
\begin{equation}
    f_{\theta, \phi}(s,a|c) = g_{\theta}(s,a|c) + \gamma h_{\phi}(s'|c) - h_{\phi}(s|c).
\label{eq:f-reward}
\end{equation}

\begin{figure}[!ht]
  \centering
  \includegraphics[width=0.85\textwidth]{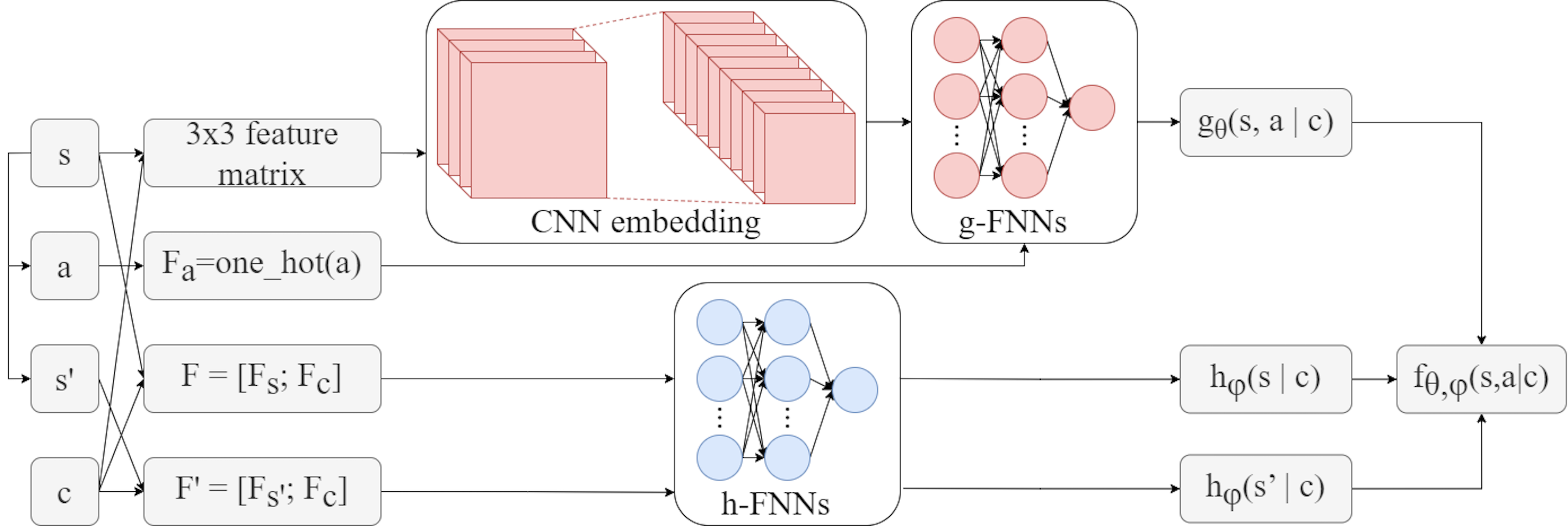}
  \caption{The network structure of the discriminator $D$}\label{fig:reward_network}
\end{figure}

To approximate $f_{\theta, \phi}(s,a|c)$, we use a network structure as displayed in Figure~\ref{fig:reward_network}. The reward estimator consists of two sub-networks, one for the reward approximator $g_{\theta}(s,a|c)$ and the other for the reward shaping term $h_{\phi}(s|c)$. For $g_{\theta}(s,a|c)$, since the reward for taking an action is generally dependent on possible next state features, we can use a two-layer CNN network to summarize features of current and adjacent states into a latent space vector similar to the policy network $G$. In addition, action features may also influence the reward. For example, travelers may prefer going straight to making right or left turns. We represent action features as a one-hot vector $F_a$, and concatenate it with the learned vector from the CNN embedding layer. $g_{\theta}(s,a|c)$ is then learned using a two-layer feed-forward network.
The shaping term $h_{\phi}(s|c)$ is also approximated using a two-layer feed-forward network, with features of current states $F=[F_s; F_{c}]$ as input. Although it is possible to use a CNN for $h_{\phi}(s|c)$ similar to $g_{\theta}(s,a|c)$, our preliminary experiments show that a feed-forward network can achieve similar performance with higher computation efficiency. This is likely because the shaping term $h_{\phi}(s|c)$ can be regarded as the expected future reward of state $s$ conditioned on the context $c$, and thus the correlation does not just depend on possible next states. With the approximated $g_{\theta}(s,a|c)$, $h_{\phi}(s|c)$ and $h_{\phi}(s'|c)$ values, we can compute $f_{\theta, \phi}(s,a|c)$ using Eq.~\eqref{eq:f-reward}. Once we have $D_{\theta, \phi}(s,a|c)$, the reward can be estimated similar to Eq.~\eqref{eq:r}:
\begin{equation}
R_{\theta, \phi}(s, a \mid c) = \log \big(D_{\theta, \phi}(s, a \mid c) \big) -\log \big(1 - D_{\theta, \phi}(s, a \mid c) \big)
\label{eq:reward}
\end{equation}

\subsubsection{Value Estimator $V$}

In model implementation, we also use a value estimator to calculate the expected return of the current state $s$ conditional on the context $c$, denoted as $V(s|c)$. The value estimated based on current state and context features $F=[F_s; F_{c}]$ using a two-layer feed-forward network. In existing model-based methods, this is usually estimated through either value iteration \citep{ziebart_maximum_2008} or solving a system of linear equations \citep{fosgerau_link_2013,mai_decomposition_2018}. However, the former lacks scalability when the number of possible combinations for $(s, c)$ becomes large, while the latter is not readily compatible with DNNs. In a model-free method like RCM-AIRL, this can be avoided by using a value estimator to directly predict $V(s \mid c)$ based on the features of $s$ and $c$. The value estimator can be trained using actual and generated trajectories.

\subsubsection{Training Algorithm}\label{method:train}

%Alg~\ref{alg:train} illustrates the training process of RCM-AIRL. 
In this section, we introduce the training process of RCM-AIRL. Recall that \cite{fu_learning_2018} showed that it would be easier to perform discrimination over state-action pairs than trajectories. Therefore, we split the training trajectories $X$ into multiple state-action-context triplets, forming a training set denoted as $\tilde{X} = \{(s^{(i)}_t, a^{(i)}_t, c^{(i)}), \forall t \in \{1, 2, ..., T_{i}\}, i \in \{1,2,... N\}$\}. During the training process, we randomly sample a batch of actual data $\tilde{X}_e$ from $\tilde{X}$ at each epoch $e$. In addition, we randomly initialize OD pairs and feed them to the policy estimator $G$ to recursively generate synthetic trajectories. Obviously, each generated trajectory ends when it reaches the destination. A generated sample data set is created by splitting the synthetic trajectories into state-action-context triplets, denoted as $\hat{X}_e$. With the actual and synthetic data $\tilde{X}_e$ and $\hat{X}_e$, the parameters of models are then updated using batch gradient descent approach. Specifically, at each epoch $e$, the reward estimator $D$ is updated by minimizing the following function:
\begin{equation}
    L_D(\theta, \phi) = - E_{\tilde{X}_e} \left[ \log \big(D_{\theta,\phi}(s, a \mid c) \big)\right] - E_{\hat{X}_e} \left[ \log \big(1 - D_{\theta,\phi}(s, a \mid c) \big)\right]
\end{equation}

The policy network is trained using a state-of-the-art policy gradient algorithm, namely Proximal Policy Optimization (PPO) \citep{schulman2017ppo}. Compared with the other approaches for policy optimization, it can effectively avoid destructively large policy updates while being much simpler to implement and tune. It works by maximizing a clipped surrogate objective:
\begin{equation}
  L_G(\pi_G) = E_{\hat{X}_e}\left[ \min\left(r_{G}(a |s, c) \hat{A}(s,a \mid c), \text{clip} \Big(r_{G}(a |s, c), 1-\epsilon, 1+\epsilon \Big)\hat{A}(s,a \mid c)\right) \right]
\end{equation}
where the term $r_{G}(a |s, c)=\pi_{G}(a |s, c) / \pi_{G, old}(a |s, c)$ denotes the probability ratio between the new policy $\pi_{G}(a |s, c)$ and the old one $\pi_{G, old}(a |s, c)$. The clip function $\text{clip}(*)$ truncates $r_{G}(a |s, c)$ within the range of $[1-\epsilon, 1+\epsilon]$ and $\epsilon$ is a pre-defined hyperparameter set as 0.2 in our case. $\hat{A}(s,a \mid c)$ denotes the estimated advantage of state-action pair $(s, a)$ conditional on the context $c$ and is calculated by adapting the generalized advantage estimator \citep{schulman_high-dimensional_2016}. Specifically, for a trajectory $x$ with length $T$, the estimated advantage is calculated as:
\begin{gather}
\hat{A}(s_t,a_t \mid c) =
    \begin{cases}
      \delta_t + \gamma \lambda\hat{A}(s_{t+1},a_{t+1} \mid c) & t = 1,2,...T_i-1,\\
      \delta_t & t = T_i,
    \end{cases} \\
\delta_t = R_{\theta, \phi}(s_t, a_t \mid c) + \gamma V(s_{t+1} \mid c) -  V(s_t \mid c)
\label{eq:gae}
\end{gather}
where the parameter $\lambda$ is used to balance bias and variance and set as 0.95, $R_{\theta, \phi}(s_t, a_t \mid c)$ is the estimated reward of the $t$-th step of the $i$-th trajectory and calculated from Eq.~\eqref{eq:reward}, $V(s_t \mid c)$ is the estimated value from the value estimator.

The value estimator is trained by minimizing the mean squared error between returns and estimated value

:
\begin{equation}
L_V(\theta_v) = E_{\hat{X}_e}\left[(Q(s,a \mid c) - V_{\theta_v}(s \mid c))^2\right],
\end{equation}
where $Q(s,a \mid c)$ denotes the return of state $s$ and action $a$ conditioned on $c$ and is calculated as the sum of $\hat{A}(s,a \mid c)$ and $V_{\theta_v, old}(s \mid c)$. $\theta_v, \theta_{v,old}$ denotes the new and old parameters of the value estimator respectively.

\subsection{Two Imitation Learning Alternatives to RCM-AIRL} \label{sec:rcm:il}
Imitation learning (IL) is closely related to IRL. While IRL tries to recover the optimal reward function $R^*$ from observed behavior, IL aims to directly learn the optimal policy $\pi^*$. IL methods can be useful, and sometimes easier to implement, when we focus mostly on route choice predictions, and are less concerned about uncovering the underlying routing preferences. In this section, we introduce two IL methods that can be adapted for route choice modeling. They will later be used as a baseline models to benchmark RCM-AIRL.

\subsubsection{RCM-BC: Behavioral Cloning for Route Choice Modeling}

In sequential decision-making domains, behavioral cloning (BC) is commonly used as a simpler alternative to IRL. BC applies supervised learning to directly train a policy that matches states to actions based on observed behavior. It is simple and often effective for small problems or with abundant data. However, like most supervised learning methods, it assumes i.i.d data, and thus does not consider the sequential dependencies between state-action pairs within each trajectory. In this section, we introduce a context-dependent BC approach to route choice modeling, named RCM-BC.

Given the observed trajectories, we divide the trajectories into independent $(s, a, c)$ triplets. The objective of RCM-BC is to learn a policy $\pi_B$ that maximizes the likelihood of the observed data:
\begin{equation}
    \max_{\pi_B} \sum_{i=1}^{N} \sum_{t=1}^{T_i} \log \big( \pi_B(a_t^{(i)} \mid s_t^{(i)}, c^{(i)}) \big)
\end{equation}

Similar to the policy network $G$ of RCM-AIRL, we approximate $\pi_B$ with a CNN as illustrated in Figure~\ref{fig:policy_network}. The output of the model is the probability distribution of actions given the state and context, and the model parameters are updated using the cross-entropy loss:
\begin{equation}
L(\theta_{B}) = -\sum_{i=1}^N \sum_{t=1}^{T_i}{\tilde{a}_t^{(i)}\log{\hat{P}(a_t^{(i)})}},
\end{equation}
where $\theta_{B}$ denotes the set of parameters in RCM-BC, $\tilde{a}_t^{(i)}$ is a one-hot vector indicating the true action at time step $t$ in the $i$-th trajectory and $\hat{P}(a_t^{(i)})$ is the corresponding probability predicted by the model.

\subsubsection{RCM-GAIL: Generative Adversarial Imitation Learning for Route Choice Modeling}

Generative adversarial imitation learning (GAIL) is a state-of-the-art model-free imitation learning algorithm \citep{ho_generative_2016}. The formulation of GAIL shares similarities with AIRL as they are both based on the adversarial learning framework. The main difference is that AIRL takes a specific form in its discriminator to recover an unshaped reward, while GAIL directly learns a policy from demonstration data and does not attempt to recover the reward function. Since GAIL has not yet been adapted to the route choice problem, in this section we introduce a context-dependent GAIL for route choice modeling, namely RCM-GAIL, which will also be used as a baseline for RCM-AIRL. 

Similar to RCM-AIRL, RCM-GAIL consists of three main modules: a discriminator,  a generator and a value estimator. The discriminator acts as a classifier to distinguish actual and generated data, while the generator aims to generate realistic trajectories that can confuse the discriminator. For fair comparison, we use the same network structure of the generator $G$ and value estimator $V$ for RCM-GAIL as those in RCM-AIRL. For the discriminator, we use the sub-network to approximate $g_{\theta}(s, a)$ in RCM-AIRL. Compared with the policy, the reward is more fundamental and thus more generalizable to unseen environments. As we will show later in Section~\ref{sec:results}), RCM-AIRL can achieve consistently better performance than RCM-GAIL for the route choice problem, albeit not by a large margin, and produce more interpretable behavioral insights. 

% \subsection{Model Interpretation}

\section{Model Evaluation Setup} \label{sec:setup}

\subsection{Data}
\begin{figure}[!ht]
  \centering
  \includegraphics[width=0.7\textwidth]{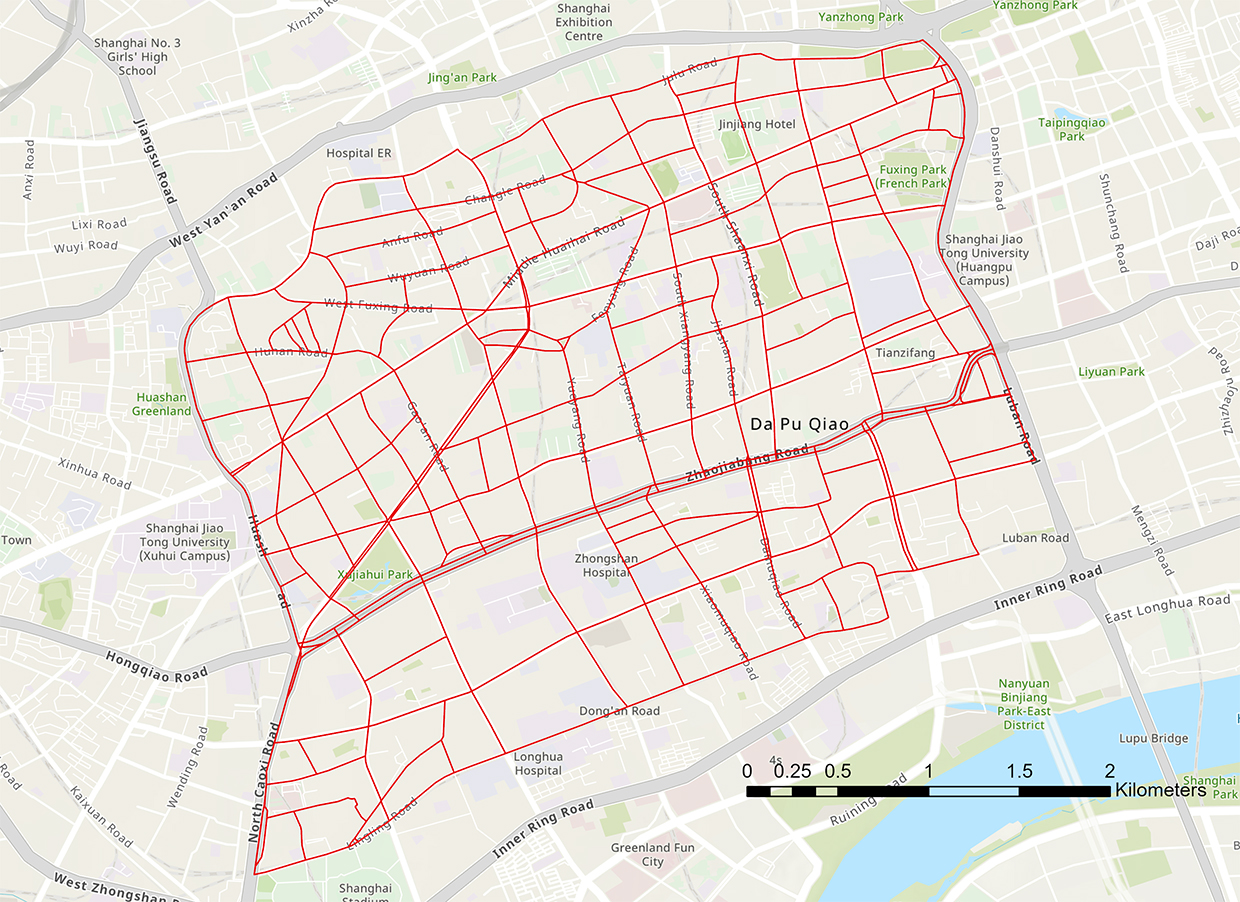}
  \caption{Selected road network in Shanghai}\label{fig:road_map}
\end{figure}

For route choice analysis, the most important data are the observed trajectories $X$. The specific dataset used in this study is retrieved from one of the major taxi companies in Shanghai, China. It contains the GPS traces of 10,609 taxis from 2015-04-16 to 2015-04-21. In our experiments, we only target a specific region of Shanghai with sufficient data coverage (see Figure~\ref{fig:road_map}). The selected road network consists of 318 intersection nodes and 712 road links, with an average link length of 199.9m. There are over 35 million records in the selected region, with an average sampling rate of roughly 10 seconds. Each record consists of the information of taxi ID, date, time, longitude, latitude and occupied flag, which is a binary indicator flagging whether the taxi is occupied. For route choice analysis, we only include occupied taxi trips with a well-defined destination to better approximate general routing scenarios \citep{liang_modeling_2022}. The GPS data has been mapped onto road networks using an open-source map matching method called \textit{Fast Map Matching} \citep{yang_fast_2018}, which is itself an implementation of the \textit{ST-Match} approach \citep{lou_map-matching_2009}. We filter out trips that are too cyclic (visiting the same link more than once). In addition, from preliminary experiments, we find that for trips less than 15 road links, around 80\% of them choose the shortest path and the difference between different models are not significant. Therefore, we only keep trips with no less than 15 road links for further experiments. This results in 24,470 trips for empirical analysis covering 664 destination links. For each trip, we keep its path trajectory (i.e., a sequence of links) as well as its contextual information (e.g., the destination, vehicle ID).

\subsection{Feature Extraction}
As introduced earlier, we consider three types of features in the model. \textbf{State features $F_s$} capture characteristics of each link. \textbf{Context features $F_c$} are features of the context, including those related to the destination. Instead of directly characterizing the destination itself, we find that it helps our model to learn faster by using features to describe the relationship between the current state and destination state. \textbf{Action features $F_a$} indicate the moving directions of the traveler. Table~\ref{table:feature} provides a detailed description of each input feature.

\begin{table}[ht!]
  \centering \footnotesize
  \caption{The description of input features}
    \begin{tabular}{p{0.25\linewidth}   p{0.65\linewidth}}
    %\addlinespace
    \toprule
    Feature & Description \\
    \midrule
    \textit{State features $F_s$} & \\
    Link length & The length of a link.\\
    Link level & Whether a link belongs to primary, secondary, tertiary, living street, residential or unclassified link level.\\
    \textit{Context features $F_c$} & \\
    Shortest distance & The shortest path distance from the current link to the destination (pre-computed using Dijkstra's algorithm). \\
    Number of links & The number of links in the shortest path from the current link to the destination.\\
    Number of turns & The number of left, right and u-turns along the shortest path from the current link to the destination.\\
    Frequency of link levels & The number of primary, secondary, tertiary, living street, residential and unclassified links along the shortest path from the current link to the destination. \\
    \textit{Action features $F_a$} & \\
    Direction & Whether the traveler is moving forward, right forward, right, right backward, backward, left backward, left or left forward.\\
    \bottomrule
    \end{tabular}%
  \label{table:feature}%
\end{table}%

\subsection{Baseline Models}

We compare RCM-AIRL, RCM-BC and RCM-GAIL against the following baseline models:

\begin{itemize} [noitemsep]
    \item \textbf{Path Size Logit (PSL)} \citep{ben-akiva_discrete_1999} is arguably the most commonly used route choice model in the literature. To account for overlaps across routes, it extends the Multinomial Logit (MNL) model by including correction terms to penalize routes that share links with other routes. The probability of choosing a route $j$ from a set of candidate routes $J$ is given by
    \begin{equation}
    P(j) = \frac{\exp(v(j) + \beta \ln (\kappa_j))}{\sum_{j' \in J}\exp(v(j') + \beta \ln (\kappa_{j'}))}
    \end{equation}
    where $P_j$ is the probability of choosing route $j$, $v(j)$ is a linear function to approximate the deterministic utility of route $j$, $\beta \geq 0$ is the path size scaling parameter, and $\kappa_j \in (0, 1]$ is the path size term for route $j$. More distinct routes with less shared links tend to have a larger path size term. In our implementation, we choose 5 shortest paths for each OD pair to form the candidate choice set based on \cite{marra_deep_2021}. Since the real trajectories are less likely to exactly match the candidate paths, for each OD training sample, we first match each trajectory in the training set to the most similar candidate path, and then the ground truth probability of candidate paths is computed as the number of matches divided by the total number of trips associated with the OD sample \citep{he_what_2020}. The path-level context features shown in Table~\ref{table:feature} are included for computing utility scores.
    \item \textbf{DNN-PSL} \citep{marra_deep_2021,he_what_2020} is an extension of PSL by using DNNs to infer the utility function of path alternatives. Compared with PSL, DNN-PSL is more flexible to capture nonlinear relationships and advanced context features. In our implementation, we employ a two-layer feed-forward network to approximate the utility function following \cite{he_what_2020}. For fair comparison, we use the same choice sets and feature settings as PSL. 
    \item \textbf{Recursive Logit} \citep{fosgerau_link_2013} is a link-based route choice model, where the route choice is modeled as a sequence of link choices using a dynamic discrete choice framework. The recursive logit gives probabilities of choosing the next state $s'$ at state $s$ according to the formula:
    \begin{equation}
    \label{eq: recursive_logit}
    P(s' \mid s) = \frac{\exp (v(s' \mid s) + V(s'))}{\sum_{{s''} \in {S'(s)}}{\exp(v(s'' \mid s) + V(s''))}},
    \end{equation}
    where $v(s'\mid s)$ is a linear function to approximate the instantaneous utility of choosing the next link $s'$ at state $s$, $V(s')$ is a value function to represent the expected downstream utility of choosing $s'$. Compared with PSL, it avoids path sampling for choice set generation and can achieve more consistent results. In our implementation, the state and action features listed in Table~\ref{table:feature} are considered for computing utility scores. While the path-level context features are not directly included, they are implicitly considered in the recursive logit as well, as they are essentially accumulated state and action features along a path. For efficient estimation, we adopt the decomposition method proposed in \cite{mai_decomposition_2018} to reduce the number of linear systems to be solved.
\end{itemize}

\subsection{Evaluation Metrics}

For model evaluation, we measure the similarity of true and predicted trajectories using the following three metrics:

\begin{itemize} [noitemsep]
    \item \textbf{Edit Distance (ED)}. Edit distance is a common way of quantifying how dissimilar two sequences are to one another by counting the minimum number of operations required to transform one sequence into the other. It is computed as: 
    \begin{equation}
    ED = \frac{1}{N}\sum_{i=1}^{N}{\min \left(\frac{Edit(\hat{x}_i, x_{i, ref})}{T_{i,ref}}, 1\right)},
    \end{equation}
    where $N$ is the number of trajectories in the test set, $\hat{x}_i$ is the $i$-th predicted trajectory, $x_{i, ref}$ is the reference trajectory, which denotes the real trajectory in the test set with the same OD, and $T_{i,ref}$ is the length of $x_{i, ref}$. It is worth noting that for an OD pair, there may exist multiple reference trajectories. In such cases, we compare the predicted trajectory with all reference trajectories in the test set and keep the best performance following \cite{choi_trajgail_2021}.
    \item \textbf{BiLingual Evaluation Understudy score (BLEU)}. In machine translation, BLEU score measures how similar a candidate text is to the reference texts, with values closer to one representing more similar texts, which can be used to measure trajectory similarities \citep{choi_trajgail_2021}. It works by comparing $n$-gram matches between each predicted trajectory to the reference trajectories with the same OD in the test set:
    \begin{gather}
    {BLEU}_n = \frac{1}{N}\sum_{i=1}^{N}{\min(1, \frac{T_i}{T_{i,ref}})(\prod_{j=1}^{n}P_j)^{\frac{1}{n}}},\\
    P_j = \frac{\sum_{m \in C_j} \min(w_m, w_{m, max})}{W},
    \end{gather}
    where $C_j$ is a set of unique $j$-gram chunks (i.e., a contiguous sub-sequence of $j$ links) found in the predicted trajectory, $w_m$ is the number of occurrences of chunk $m$ in the predicted trajectory; $w_{m,max}$ is the maximum number of occurrences of chunk $m$ in one reference trajectory, and $W$ is the total number of chunks in the predicted trajectory. %We use $n=4$ in our case.
    \item \textbf{Jensen-Shannon Distance (JSD)}. In probability theory, JSD measures the similarity between two probability distributions based on KL (Kullback–Leibler) divergence. In our case, we use the route frequencies (i.e., the occurrence of each unique route divided by the total number of trajectories) to represent the probability distribution of a trajectory dataset. Predicted trajectories that do not exist in the test set are labeled as ``unseen'' routes following \cite{choi_trajgail_2021}. Given the probability distribution of the observed and predicted trajectories $p$ and $q$, the JS distance is defined as follows:
    \begin{equation}
        d_{js} = \sqrt{\left(D_{KL} \big(p \parallel \frac{p+q}{2} \big) + D_{KL} \big(q \parallel \frac{p+q}{2} \big) \right)/2}
    \end{equation}
    where $D_{KL}$ is the KL divergence, and $D_{KL} (p \parallel q)$ is also known as the relative entropy of $p$ with respect to $q$, which is defined as:
    \begin{equation}
        D_{KL} (p \parallel q) = \sum_i p_i \log \frac{p_i}{q_i}.
    \end{equation}
    \item \textbf{Log Probability (LP)}. For link-based route choice models, a common evaluation metric is the average log probability of real trajectories in the test set under the given model \citep{ziebart_maximum_2008}, which is given as:
    \begin{equation}
    LP = \frac{1}{N}\sum_{i=1}^{N}{\sum_{j=1}^{L_i-1}log(\hat{P}(s_{j+1}|s_j,c))},
    \end{equation}
    where $\hat{P}(s_{j+1}|s_j,c)$ is the estimated probability of moving to state $s_{j+1}$ from state $s_j$ with context $c$, $L_i$ is the length of the $i$-th trajectory in the test set. It is worth noting that for path-based models, LP does not exist since the real paths in the test set might not exist in the candidate paths.
\end{itemize}

\subsection{Experiment Settings}

We use cross validation to compare the performance of different models. Specifically, the original dataset is randomly partitioned into 5 equal sized subsets, each with 4894 trips. Each subset is used once as the test data, and the training data is sampled from the remaining 4 subsets. Since DNNs have long been criticized for their need for large amounts of training data, we vary the training data size from 100, 1000 to 10000 trajectories to investigate how the amount of training data influences the model performance. All the experiments are repeated three times to consider model instability. 

For the training of RCM-AIRL, through experiments, we set the number of iterations as 1000, 2000 and 3000 for the training data size of 100, 1000 and 10000 trips respectively. More details on the hyperparameters used in RCM-AIRL is shown in Table~\ref{table:hyperparameter}. Specifically in our case, we find that the design of context-dependent rewards are crucial for the successful training of our proposed model. As mentioned before, the standard AIRL model can get stuck in loops and never reach the destination. However, by incorporating context features, especially those related to the destination, RCM-AIRL can generally achieve satisfying results in our experiments. This is potentially because, compared with state-action features, the information provided by destination-related features can better guide the agent to focus on reasonable actions and make meaningful progress to the destination, thus reducing the difficulty of model training. This suggests that a careful selection of input features is critical for training model-free IRL for route choice modeling. Another technique we find useful is to provide a sufficient number of generated samples per iteration for model training. In our preliminary experiments, we vary the number of samples from 2048, 4096 to 8192 and find that increasing the number of samples can help reduce the prediction error. Previous research has shown that GAN models may result in ``mode collapse" when the number of generated samples is small \cite{choi_trajgail_2021}, resulting in a limited variety of generated data. Using a sufficient number of generated samples at each training iteration can solve this problem in our case.

\begin{table}[ht!]
  \centering \footnotesize
  \caption{Hyperparameters used for RCM-AIRL}
    \begin{tabular}{lc}
    %\addlinespace
    \toprule
    Hyperparameter & Value \\
    \midrule
    Number of dicriminator updates per iteration & 1\\
    Number of PPO updates per iteration & 10\\
    Batch size for PPO updates & 64 \\
    Learning rate & 3e-4\\
    Discount rate of reward ($\gamma$) & 0.99 \\
    Number of generated samples per iteration & 8192 \\
    Number of CNN layers & 2\\
    Output channel and kernel size in the first CNN layer & 20, 3\\
    Output channel and kernel size in the second CNN layer & 30, 2\\
    \bottomrule
    \end{tabular}%
  \label{table:hyperparameter}%
\end{table}%

\section{Results} \label{sec:results}

\subsection{Model Comparison for Prediction Performance} \label{sec:performance}

\begin{figure}[!ht]
  \centering
  \includegraphics[width=\textwidth]{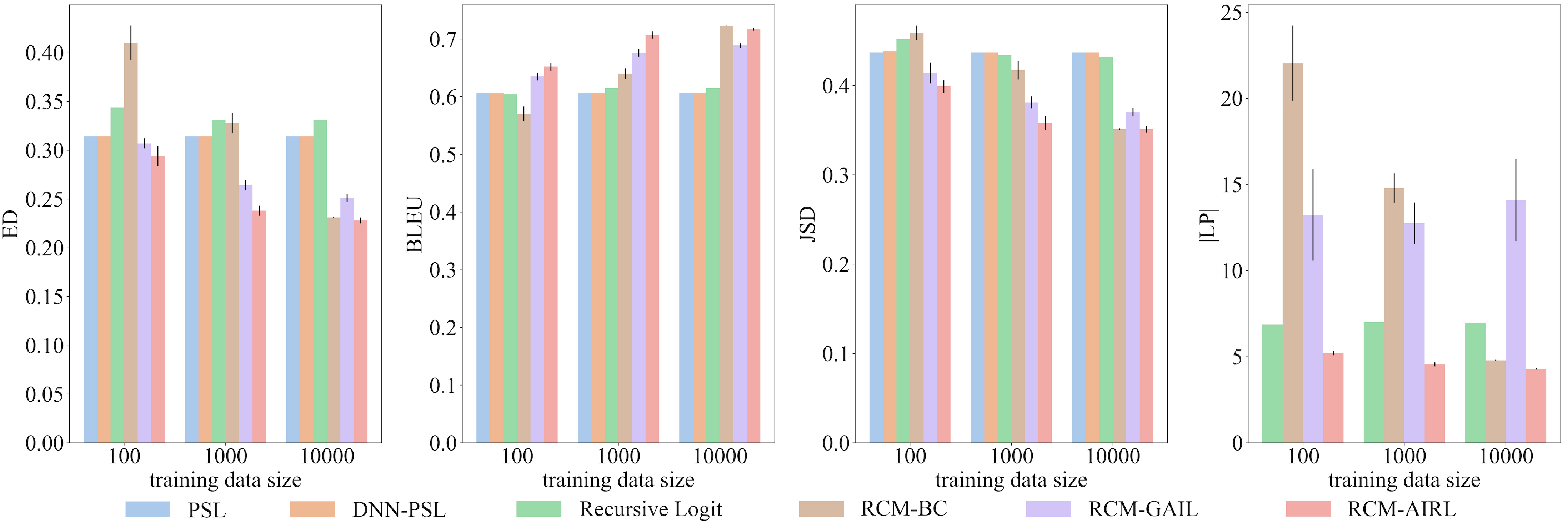}
  \caption{Route choice prediction performance of different models. The results are averaged over 5-fold cross validation and the error bars represent the standard deviations over 3 repeated experiments.}\label{fig:performance}
\end{figure}

In this section, we evaluate the performance of different models over 5-fold cross validation and the average performance of different models is summarized in Figure~\ref{fig:performance}. We use the $t$-test to evaluate the performance difference between different models. It is found that RCM-BC, RCM-GAIL and RCM-AIRL all perform significantly better than the existing baseline models with $p$-value small than 0.001, suggesting the effectiveness of deep IRL/IL methods for route choice modeling. Compared to RCM-BC and RCM-GAIL, RCM-AIRL performs consistently better regarding all evaluation metrics, confirming the advantage of the IRL framework over IL for route choice modeling. Unlike RCM-BC, RCM-AIRL can better estimate long-term reward and capture the probability distribution of route choices. Relative to RCM-GAIL, RCM-AIRL can better recover the underlying reward function, which makes it more robust for route choice modeling under different routing environments.

Among baseline models, PSL and DNN-PSL achieve similar performance, suggesting that the contribution of DNNs on PSL might not be significant on our dataset, though this might change when more sophisticated features are included. On the one hand, recursive logit performs slightly worse than PSL in terms of ED/BLEU/JSD, especially when the training data size is small. This is likely because recursive logit assigns a fixed immediate reward (i.e., $v(s'|s)$ in Eq.~\eqref{eq: recursive_logit}) for each state-action pair regardless of the destination, making it harder to train when there are many destinations.  On the other hand, recursive logit can produce reasonably good LP, when path-based models cannot. This is as expected since the former considers all possible path choices, while the latter are limited by the choice set generation algorithms.

To account for potential uncertainty of the model results, we compute the standard deviation of the three repeated experiments for each cross validation process and the averaged standard deviations over 5-fold cross validation are shown as the error bars in Figure~\ref{fig:performance}. It can be found that PSL and recursive logit have standard deviations of zero, which is as expected. This can be explained by the consistency property of the maximum likelihood estimator and the fact that logit models with linear utility functions generally guarantee the convergence to global optima. In our case, the standard deviation of DNN-PSL is also zero, which might be because path-based models are easy to optimize. The results of RCM-BC are quite unstable with a small training set, and its stability can be improved by enlarging the training set. For RCM-GAIL, while it achieves stable results for trajectory-level metrics (i.e., ED, BLEU and JSD), its predicted log probability has large variance regardless of training data size, suggesting that GAIL might be relatively poor in achieving a stable probabilistic prediction in our case. Compared with RCM-BC and RCM-GAIL, RCM-AIRL is more stable for all evaluation metrics based on different training data. This verifies that RCM-AIRL can provide stable prediction results even with a limited number of training data.

\begin{figure}[!ht]
  \centering
  \includegraphics[width=\textwidth]{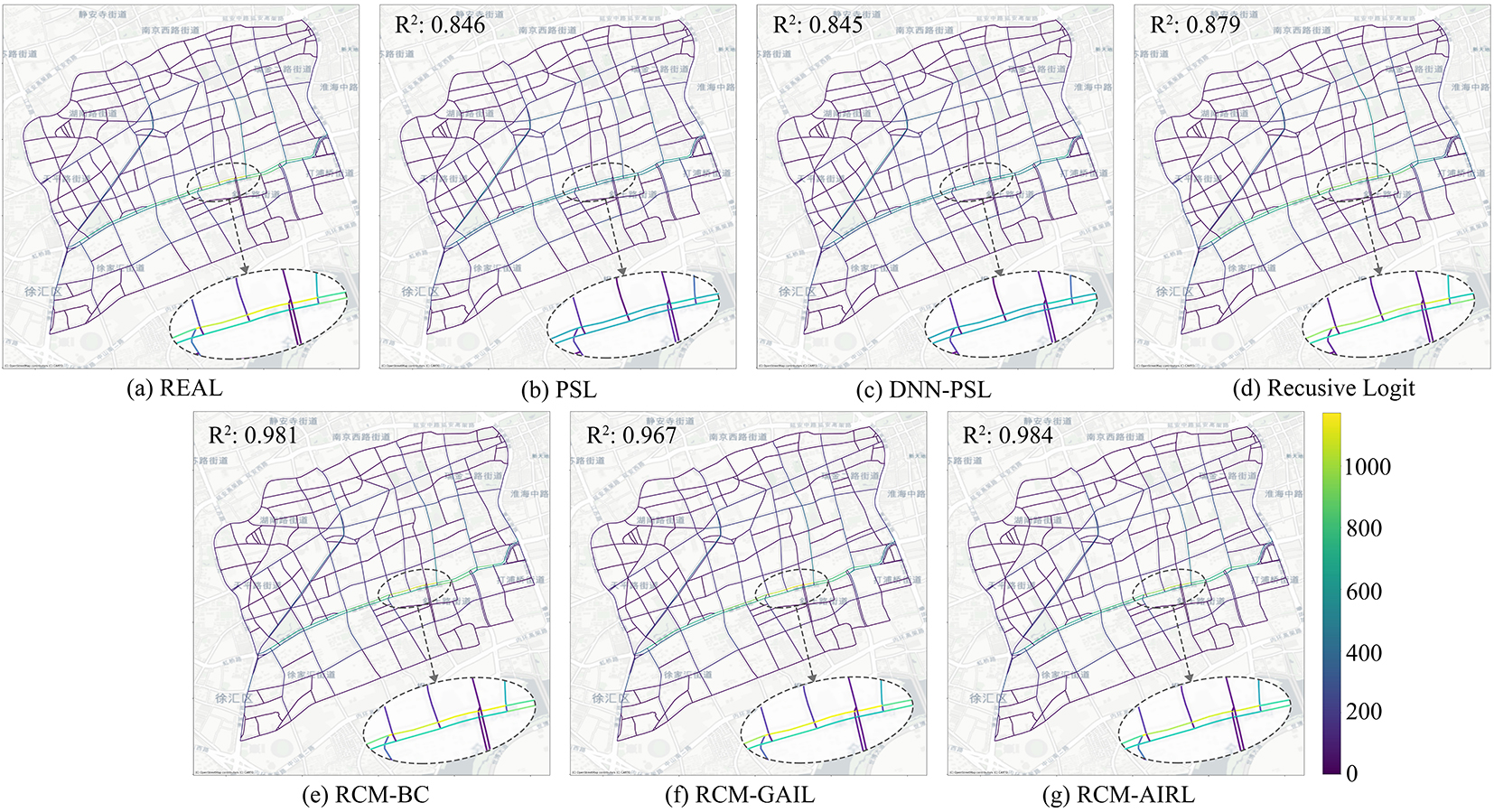}
  \caption{Predicted link flow distribution using different models}\label{fig:flow}
\end{figure}

In transportation planning, route choice models are often used for network traffic flow estimation. Here we also demonstrate the capability of our proposed models for network flow estimation. Using the proposed IL/IRL models, the context-dependent link choice probabilities can be computed as outputs of the learned policy network. Given an OD matrix characterizing the travel demand between each OD pair on the network, we can then compute the expected network flow by adopting a simulation-based approach \citep{zimmermann_bike_2017}. Specifically, we first draw the same number $r$ of paths for each OD pair based on the learned link choice probabilities. The path choice probabilities are known for each of these paths and we normalize them so that the sum over the $r$ simulated paths for each OD equals one. For trip assignments, we distribute the trips given by the observed OD matrix according to the computed path probabilities. In our case, we set $r$ as 5 for fair comparison with path-based models. It is worth noting that path-based models generate candidate paths using pre-defined choice set generation algorithms while link-based models can simulate candidate paths according to the learned policies without pre-defined choice sets. The computation results of the expected network traffic flow using different models are shown in Figure~\ref{fig:flow}. For each model, the figure also shows the corresponding $R^2$ (averaged across 5-fold cross validation) for link-level flow prediction (compared to the actual link flows). RCM-AIRL is found to achieve the highest $R^2$, followed closely by RCM-BC, validating the advantage of the proposed methodology.

\begin{figure}[!ht]
  \centering
  \includegraphics[width=\textwidth]{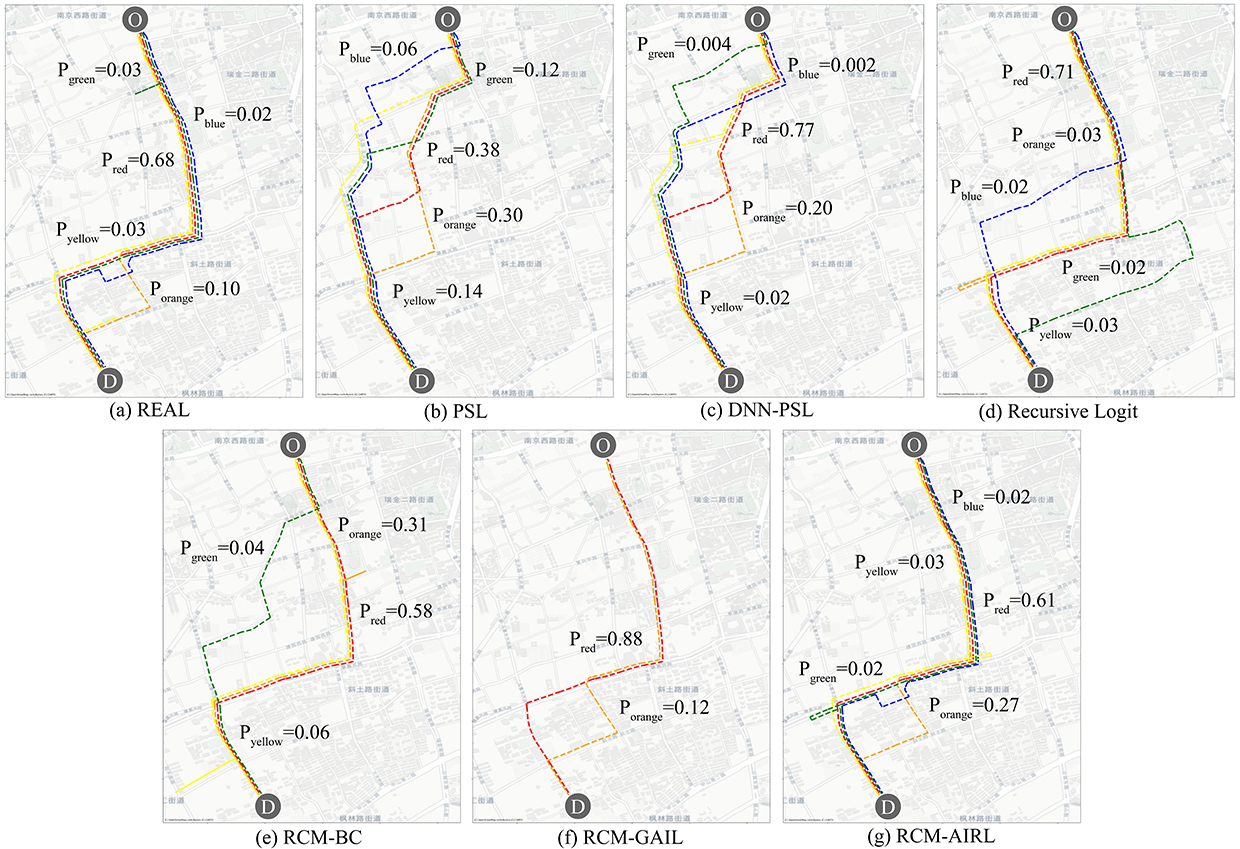}
  \caption{Example route choices predicted by different models for an example OD pair}\label{fig:visualize}
\end{figure}

To gain more intuition about model outputs, Figure~\ref{fig:visualize} shows the actual and predicted route choices for an example OD pair. In our dataset, 59 actual trajectories are observed for this OD, distributed across 12 unique paths. For clarity, we only display the top 5 routes and their relative proportions in Figure~\ref{fig:visualize}(a). The predictions are generated using different models trained on 10000 trips. For comparison, 59 trajectories are generated based on each model's predicted path probabilities. Due to the restriction of choice set generation, only the 5 shortest paths are evaluated for the path-based models, as shown in Figure~\ref{fig:visualize}(b-c). For link-based models, however, any number of unique paths ($\leq 59$) can be predicted, but we only keep at most 5 distinct paths for visualization in Figure~\ref{fig:visualize}(d-g). It is found that RCM-AIRL can well capture the distribution of paths in the observed trajectories, in terms of both the concentration and variability. In \ref{appendix:b}, we also show how the predicted route choices can change when one link along the actual trajectories is removed, demonstrating the model's ability to work under hypothetical network change situations.

\subsection{Model Comparison for Computational Efficiency}

In this section, we compare the computational efficiency of different models and the results are displayed in Table~\ref{table:computation}. All the experiments are implemented on a server with i9-10900KF CPU with 256G RAM. The computation time is broken down into three parts, namely feature engineering (FE), model optimization (MO), and path generation (PG). The first two are parts of model training, while the last part is for model testing. It is shown that PSL and DNN-PSL are most efficient for MO regardless of the size of training data. However, they are much slower than the other models for PG. This is because path-based models need additional time for choice set generation for each OD pair, which can be computationally expensive. Among link-based models, recursive logit is the most efficient for training, because it does not require any path-level features and is relatively efficient to estimate thanks to the decomposition method that allows for one system of linear equations to be solved for all destinations \citep{mai_decomposition_2018}. Deep learning models generally have lower training efficiency but higher testing (PG) efficiency. Compared with RCM-GAIL and RCM-AIRL, RCM-BC is more efficient for both training and testing due to its simple model architecture. Different from the other models, the training time of RCM-GAIL and RCM-AIRL is mainly dependent on hyperparemeter settings including the number of iterations and the number of samples per iteration. Although the training efficiency of RCM-AIRL is relatively poor compared with the other models, especially with small training data, we argue that the model can be trained once for repeated use. In practice, route choice models are often used for large-scale deployment as part of network traffic flow prediction/simulation. In such cases, PG is more important than FE or MO, and the overall computational efficiency of the proposed model should be competitive. 

\begin{table}[ht!]
  \centering \footnotesize
  \caption{Computation time for feature engineering (FE), model optimization (MO), and path generation (PG) of different models. The results are averaged over 5-fold cross validation.}
    \resizebox{\linewidth}{!}{%
    \begin{tabular}{cccccccccccc}
    %\addlinespace
    \toprule
    \multicolumn{2}{c}{\multirow{2}{*}{Models}} & \multicolumn{3}{c}{100 training trips} & \multicolumn{3}{c}{1000 training trips} & \multicolumn{3}{c}{10000 training trips}\\
    & & FE(s) & MO(s) & PG(s) & FE(s) & MO(s) & PG(s) & FE(s) & MO(s) & PG(s) \\
    \midrule
    \multirow{2}{*}{Path-based} & PSL & 28.8 & 0.3 & 1370.5 & 277.8 & 1.8 & 1370.5 & 3166.7 & 12.6 & 1370.4 \\
    & DNN-PSL & 28.8 & 0.8 & 1370.4 & 277.8 & 5.7 & 1370.4 & 3166.7 & 42.3 & 1370.4 \\
    \multirow{5}{*}{Link-based} & Recursive Logit & 0.0 & 22.2 & 4.0 & 0.0 & 44.4 & 4.0 & 0.0 & 238.8 & 4.1\\
    & RCM-BC & 3840.0 & 124.8 & 1.7 & 3840.0 & 1174.1 & 1.6 & 3840.0 & 11487.4 & 1.6 \\
    & RCM-GAIL & 3840.0 & 6112.8 & 1.7 & 3840.0 & 9305.3 & 1.6 & 3840.0 & 18270.9 & 1.6 \\
    & RCM-AIRL & 3840.0 & 5990.4 & 1.6 & 3840.0 & 8962.5 & 1.5 & 3840.0 & 18307.0 & 1.5 \\
    \bottomrule
    \end{tabular}%
    }
  \label{table:computation}%
\end{table}%

\subsection{Model Generalizability to Unseen Destinations} \label{sec:generalizability}

Compared to simpler linear models, DNNs are more likely to overfit and may easily suffer from low generalizabily when applied to testing data different from training data. For route choice analysis, this may occur when the model encounters new destinations that are unseen during model training. To investigate model generalizability, we design specific training and testing datasets so that destinations in the test set are never seen during the training phase. This setting allows us to properly evaluate how the model generalize to new routing scenarios. 

\begin{figure}[!ht]
  \centering
  \includegraphics[width=\textwidth]{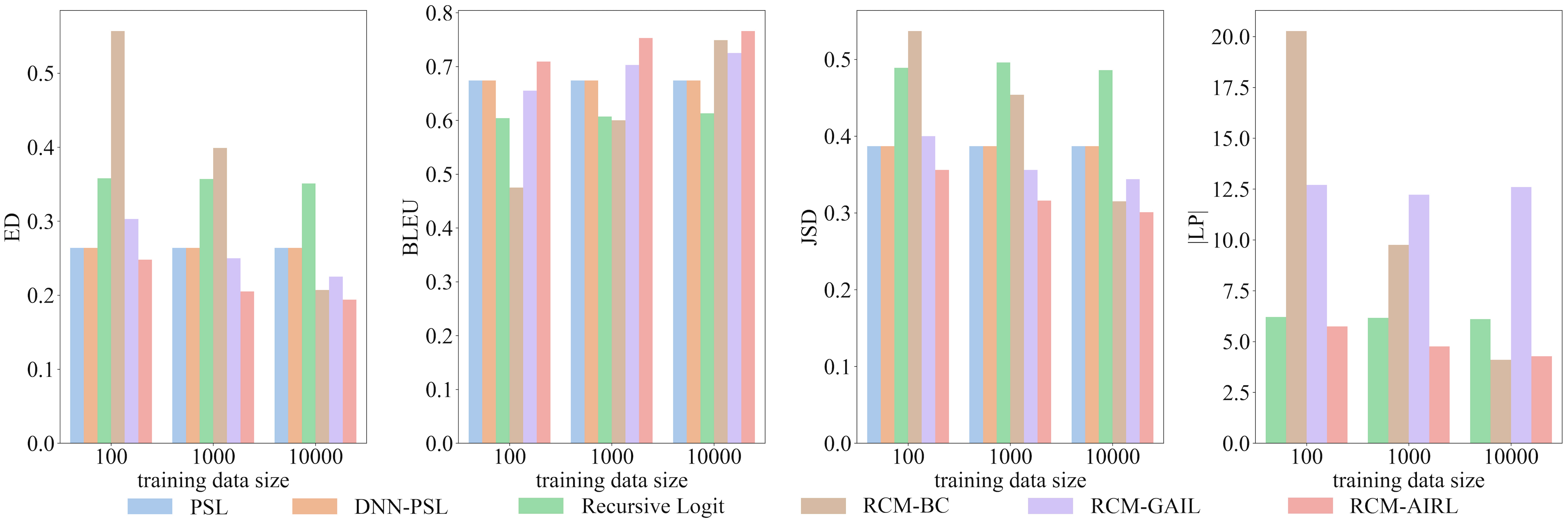}
  \caption{Route choice prediction performance of different models for unseen destinations}\label{fig:generalization}
\end{figure}

Figure~\ref{fig:generalization} summarizes the prediction performance of different models on unseen destinations using different amounts of training data. It is found that RCM-AIRL still achieves the best performance compared with the other models. With only 100 training trajectories, RCM-AIRL can already outperform traditional models. As the training data size increases to 1000 trajectories, the relatively improvement is enlarged. The results with 10,000 trajectories are slightly better than the results with 1000 trajectories. This suggests that RCM-AIRL is generalizable to unseen data and has relatively low requirement of training data. The performance of PSL and recursive logit does not change much as the training data increases. This is reasonable since these models assume linear relationships and have only a small number of parameters to estimate. Adding training data does not contribute much to the performance of DNN-PSL either. This is potentially because DNNs are restricted by the pre-defined choice sets in path-based models and cannot fully uncover the complex relationships. Unlike conventional DCMs or path-based deep learning models, the performance of RCM-BC is significantly affected by the amount of training data. With 100 or 1000 training trips, RCM-BC performs much poorer than the other models, while with 10000 training trips, RCM-BC can achieve similar performance with RCM-AIRL. This is because behavior cloning assumes i.i.d data and thus has a high requirement of training data to uncover meaningful relationships. With 100 training trips, RCM-GAIL is unable to provide satisfactory results either. With more training data, the performance of RCM-GAIL can be improved greatly, but is still worse than RCM-AIRL. This suggests that the training of IL methods tend to require more data than comparative IRL methods, at least for route choice modeling. 

\subsection{Interpretability of Learned Route Choice Behavior}

In addition to modeling realistic route choice behaviors, understanding the underlying human routing preferences is important for transportation network planning, policy design, and infrastructure investment. Interpretability of such models are also essential for ensuring model trustworthiness. In this section, we explore why RCM-AIRL makes such predictions from both global and local perspectives.

\begin{figure}[!ht]
  \centering
  \includegraphics[width=\textwidth]{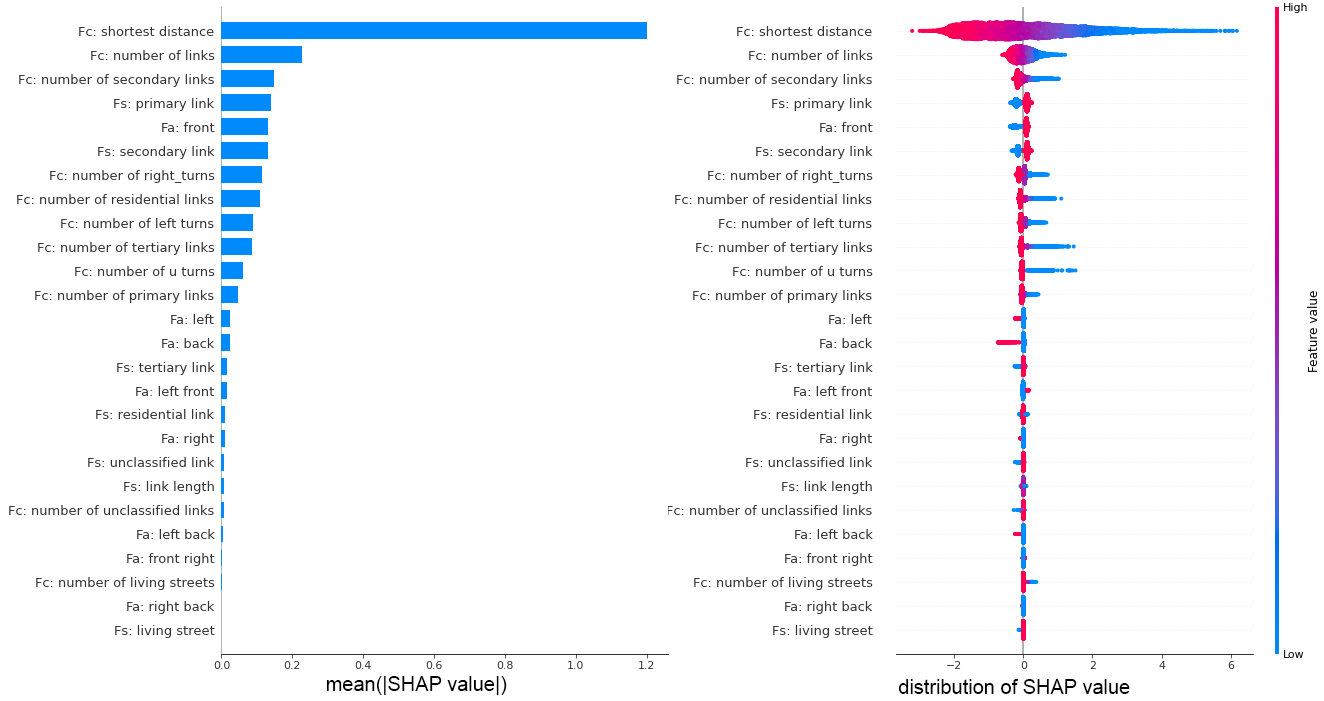}
  \caption{Distribution of Shapley values for all features in the discriminator of RCM-AIRL. ($F_c$ denotes context features, $F_s$ denotes state features and $F_a$ denotes action features)}\label{fig:shap}
\end{figure}

From a global perspective, we use SHapley Additive exPlanations (SHAP) to understand the effects of input features influence based on the reward network of RCM-AIRL. SHAP is an explainable AI technique that uses a game theoretic approach to explaining machine learning models \citep{lundberg_unified_2017}. Specifically, it assigns each feature an optimal Shapley value, which indicates how the presence or absence of a feature changes the model prediction result (i.e., the estimated reward in our case). Recall that the input of the reward network consists of three types of features: state features of the current link, action features denoting the movement direction, and context features describing the trip characteristics. We show the distribution of SHAP values for different features in Figure~\ref{fig:shap}. It is apparent that the mean Shapley value of the distance to the destination is much larger than the other features. From the distribution of SHAP values, we can further find that shorter distances are associated with higher SHAP values. As expected, travelers prefer shorter routes to reduce their travel costs. In addition to travel distance, the link level also influences route choices: primary and secondary links are related with positive SHAP values, indicating that travelers prefer main road links, which usually have better driving conditions and allow vehicles to travel faster. Another factor that influences driving behavior is the movement direction. It is found that moving forward (e.g., front) is assigned with a positive SHAP value, while moving left, right and backward are assigned with negative SHAP values, suggesting that drivers dislike turns. Moreover, left turns are associated with higher feature importance than right turns. In cities with right-hand traffic (e.g., Shanghai), making left turns usually requires longer waiting time and more careful maneuver at intersections, and thus derives higher cost compared with right turns. 

It is worth noting that several of the top features in Figure~\ref{fig:shap} are destination-related context features. Compared to link features, these destination-related features are found to be more important in the model-free settings. Unlike existing link-based models, RCM-AIRL uses the generator to sample series of state-action pairs that form realistic trajectories to the destination. These features encourages the model to explore actions that are more likely to move the agent toward the destination. Without them, the model would depend solely on state-action features and may have troubles reaching the destination efficiently. In \ref{appendix:c}, we show that the model performance can be much worse without these destination-related features. Similar interpretability analysis is also performed for other baseline models for comparison, and the results can be found in \ref{appendix:d}.

\begin{figure}[!ht]
  \centering
  \includegraphics[width=\textwidth]{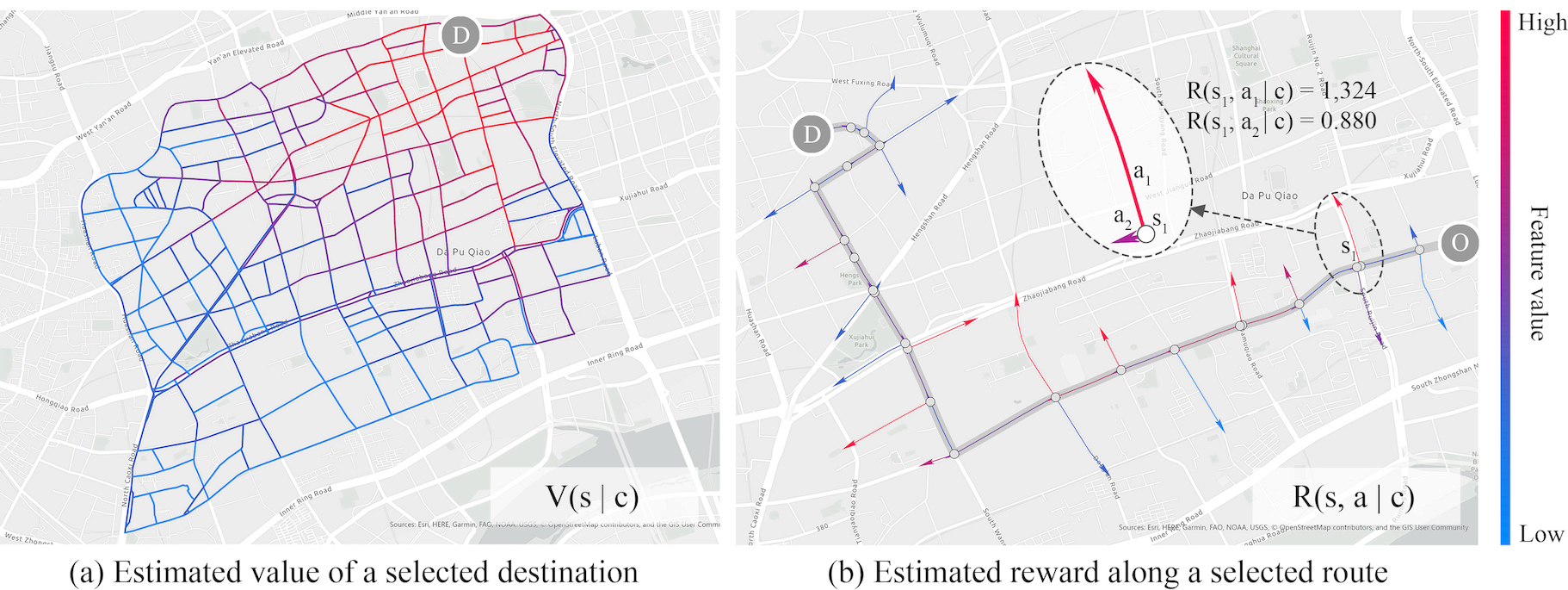}
  \caption{Examples of estimated $V(s |c)$ and $R(s, a | c)$ from RCM-AIRL}\label{fig:local_interpretation}
\end{figure}

From a local perspective, we can use the model estimates to quantitatively compare the reward scores assigned to different states and actions. Specifically, the reward $R(s, a \mid c)$ learned from the reward estimator can be used to approximate the instantaneous utility derived from choosing an action $a$ at state $s$ conditioned on context $c$, while the value $V(s \mid c)$ learned from the value estimator can be used to represent the expected downstream utility of state $s$ with context $c$. In Figure~\ref{fig:local_interpretation}(a), we visualize the estimated values of different links to a specific destination. Generally, links that are closer to the destination would be assigned with a higher value score. In this way, the model is encouraged to choose the next link associated with a shorter distance to the destination. In Figure~\ref{fig:local_interpretation}(b), we illustrate the estimated rewards of taking different actions along a selected route. It can be found that drivers do not always choose the action with the highest immediate reward. For example, at state $s_1$, the traveler chooses $a_2$, even when $R(s_1, a_2 \mid c) < R(s_1, a_1 \mid c)$. This indicates that travelers consider not only short-term but also long-term rewards/utilities, demonstrating the effectiveness of IRL in considering the cumulative reward through out the whole trip. RCM-AIRL, like any IRL methods can account for expected cumulative rewards in the future, while a standard supervised learning method like RCM-BC assumes i.i.d. data and only focuses on the instantaneous return.

\subsection{Incorporating Latent Individual Features}
Classical DCMs for choice modeling often include both attributes of the alternatives (i.e., paths or links) and characteristics of decision makers (i.e., individual travelers) such as age, gender, income, etc. In our proposed modeling framework, characteristics of each individual traveler can be treated as part of the context information $c$. Due to data limitations, we are not able to incorporate explicit individual characteristics in the experiments. Nevertheless, because each GPS trace is associated with a vehicle ID (assumed to be an individual identifier), it is possible to extract latent individual features directly from data with DNNs. Specifically, we can learn an embedding vector for each individual traveler as part of the route choice model, which would enable personalized routing behavior prediction that is useful for some transportation applications \citep{liu_personalized_2022}. 

In this section, we conduct experiments to explore whether the latent individual features learned from DNNs can add to model prediction performance. To ensure that abundant training data for each driver is provided, we select 50 drivers with the most trips for the experiments. For each driver, we split 80\% of trajectories into training data and 20\% into test data. The results with and without individual embedding features are summarized in Table~\ref{table:individual}. It is found that incorporating individual features can indeed lead to a small increase in prediction performance when sufficient training data is provided. This also validates the advantage of DNNs in incorporating complex and high-dimensional context features. When training data is sufficient, it is possible to include more context features, such as time of the day and weather, which we will leave for future work. When only a limited amount of training data is available, it is still possible to train a powerful model with good generalizablity and interpretability using handcrafted network features, as illustrated in our previous experiments.

\begin{table}[ht!]
  \centering \footnotesize
  \caption{Performance comparison of incorporating embedded individual features or not using RCM-AIRL}
    \begin{tabular}{ccccc}
    %\addlinespace
    \toprule
   With/without individual embedding & ED & BLEU & JSD & LP\\
    \midrule
    Without individual embedding & 0.245 & 0.700 & 0.391 & -4.206 \\
    With individual embedding & \textit{\underline{0.223}} &\textit{\underline{0.716}} & \textit{\underline{0.380}} & \textit{\underline{-3.869}}\\
    \bottomrule
    \end{tabular}%
  \label{table:individual}%
\end{table}%

\section{Discussion} \label{sec:conclusion}
This study proposes a deep inverse reinforcement learning (IRL) framework for route choice modeling. In this framework, the route choice problem is formulated as a Markov Decision Process, and the goal is to recover the underlying reward function (i.e., routing preferences) that can best explain, and ultimately predict, actual human route choice behavior. Both the reward and policy functions are assumed to be context-dependent and can be approximated with DNNs, making the framework more generalizable to diverse routing scenarios with many destinations and heterogeneous agents. Specifically, a context-dependent adversarial IRL (AIRL) approach is adapted for route choice modeling, so that the underlying reward and policy functions can be learned efficiently from observed trajectories in a model-free fashion. To validate the proposed model (RCM-AIRL), extensive experiments are conducted using taxi GPS data from Shanghai, and results confirm its improved performance over conventional discrete choice models as well as two imitation learning baselines---behavioral cloning (RCM-BC) and Generative Adversarial Imitation Learning (RCM-GAIL). The performance improvement holds even for unseen destinations and limited training data. We also demonstrate the model interpretability using shapely additive explanations, which reveal the reward distribution across the road network and the effects of different features. The proposed methodology is general and should be adaptable to other route choice problems across different modes and networks.

The proposed RCM-AIRL is distinct from other baseline models, and thus provides a new and promising direction for future development of route choice models. Unlike widely used path-based models such as Path Size Logit, RCM-AIRL is link-based and does not require path sampling. Therefore, it can leverage detailed network features and potentially deal with dynamic routing environments. Compared to existing link-based models like recursive logit, the inclusion of deep architectures in RCM-AIRL makes it flexible enough to incorporate diverse features (of the state, action, and trip context) and capture their complex relationships. The two imitation learning alternatives, RCM-BC and RCM-GAIL, share some similarities with RCM-AIRL but show relatively worse performances. Unlike IRL, imitation learning does not aim to recover the reward function and instead tries to directly estimate the policy. However, the reward function (i.e., routing preferences) is more fundamental, robust and generalizable than the policy (i.e., routing patterns). In addition, the ability to uncover underlying routing preferences is important for model trustworthiness and policy implications. Between the RCM-BC and RCM-GAIL, the former assumes i.i.d data and essentially ignores the sequential dependencies across link choices within the same trip. Nevertheless, behavioral cloning is structurally simple, easier to train, and can achieve good performance with abundant training data, making it a useful alternative in certain situations.

As mentioned previously, the proposed RCM-AIRL is an example of model-free method for link-based route choice modeling. This is quite different from existing model-based methods, which learn the reward/utility functions through either value iteration or solving a system of linear equations. In model-based methods, all possible trajectories need to be evaluated, even though most of them are unrealistic and unlikely to be chosen by human travelers. Generally, model-free methods are more flexible and computationally simpler, and needs no accurate representation of the environment (such as the road network and traffic rules) in order to be effective \citep{sutton_reinforcement_2018}. However, they do come with certain disadvantages as well, including their reliance on trial-and-error learning. For route choice modeling, it can be challenging to sample realistic trajectories to a specific destination without proper model design and feature engineering. In the case of RCM-AIRL, we find that destination-related features play an important role to encourage efficient exploration and improve learning efficiency. Note that we do not claim model-free methods to be superior to model-based ones. Rather, some combination of the two could be most appealing. Actually, RCM-AIRL, while largely model-free, still has some model-based characteristics. In addition to the use of destination-related features computed based on shortest path searching, it also considers the dependencies between adjacent links in the road network through CNNs (in the policy and reward estimators). Both model design choices leverage the known road network structure for efficient model learning, and they are indeed shown to improve prediction performance. A promising research direction is to explore other ways to combine model-based and model-free IRL methods (enhanced with DNNs) for more efficient and robust route choice modeling.

Future research can extend this work in several directions. First, features tested in this study are mostly based on prior literature and can be further extended. One advantage of deep learning methods is their ability to incorporate high-dimensional features. Therefore, future studies may examine the effects of other complex features, such as traffic dynamics, road network design and land use patterns, on human routing behavior. Second, the experiments in this study are only based on taxi GPS data on a relatively small road network. Future studies should explore how to extend the proposed model to a larger network and compare performance across multiple travel modes. The former requires additional measures to tackle data sparsity problems and the latter needs inclusion of mode-specific features. Third, in this work we only consider model generalizability to unseen destinations in the same network. Future research may further investigate how such deep learning models can generalize to new networks and entirely new cities. Again, this will depend on the specific network representation and feature selection in the model. Last, this study only considers route choice modeling when the trip destination is given. In real-time intelligent transportation systems (ITS) applications, we often do not know the final destinations of moving agents on the road network. Future research may try to develop multitask methods to simultaneously predict the trip destination and route choice.

\section*{Acknowledgements}
This research is supported by the Seed Funding for Strategic Interdisciplinary
Research Scheme, The University of Hong Kong (URC102010057). It also used high performance computing facilities offered by Information Technology Services, the University of Hong Kong. The authors would like to thank the anonymous reviewers for their detailed and constructive comments, and Zhengyi Ge for assisting in the paper revision.

%\section*{Declaration of Interests}
%None.

\appendix

\section{Definition of action space}\label{appendix:a}

To define the action space for link-based route choice modeling, a series of decisions need to be made, including the following:
\begin{itemize}
    \item \textbf{Link IDs vs movement directions}. The action space can be defined based on specific link IDs or movement directions (i.e., discretized link-to-link turning angles). We choose the latter for three reasons. Firstly, it is a more efficient representation of the action space since the number of distinct directions are much less than the number of link IDs. Secondly, it enables the model to learn direction-specific routing preferences, because the rewards related to the similar direction change are likely to be correlated. Defining direction-based actions enables the model to learn such correlations from the data. Finally, it also more realistically reflects routing behavior. When people navigate through a road network, they naturally make decisions based on turns instead of specific link IDs.
    \item \textbf{Number of directions}. It is computationally convenient to discretize the continuous turning angle degrees to a finite number of directions (e.g., forward, left, backward, and right). A higher number of directions means that each of them covers a narrower range of headings, leading to more precise directional representation at the cost of increased model complexity. In this study, we choose 8 directions---forward (F), forward left (FL), left (L), backward left (BL), backward (B), backward right (BR), right (R), and forward right (FR), as shown in Figure~\ref{fig:action_map}. This is based on our prior study \citep{liang_nettraj_2022}, which showed that 8 directions can uniquely map most link-to-link turning angles in Shanghai's road network. An additional advantage of having 8 directions is that they can be conveniently organized into a $3\times 3$ matrix, which makes it straightforward to apply convolution calculation, as shown in Figure~\ref{fig:policy_network}. 
    \item \textbf{Global vs local action space}. While 8 directions are adequate in mapping most link-to-link turning angles, in reality most links are only connected to 2-3 other outgoing links. Also, the number of outgoing links varies across links. To further improve the model efficiency, we can adapt the the global action space $A$ to local network layout. In this study, we construct a link-specific local action space $A(s)$ based on the network layout, and use it to adapt the action space based on the current state $s$. If the agent is at a link with only 3 outgoing links, $A(s)$ would only allow 3 of the 8 directions to be chosen. This ensures that the route choices predicted by the model would always be valid.
\end{itemize}

\section{Counterfactual analysis: A link closure example}\label{appendix:b}
\begin{figure}[!ht]
  \centering
  \includegraphics[width=0.75\textwidth]{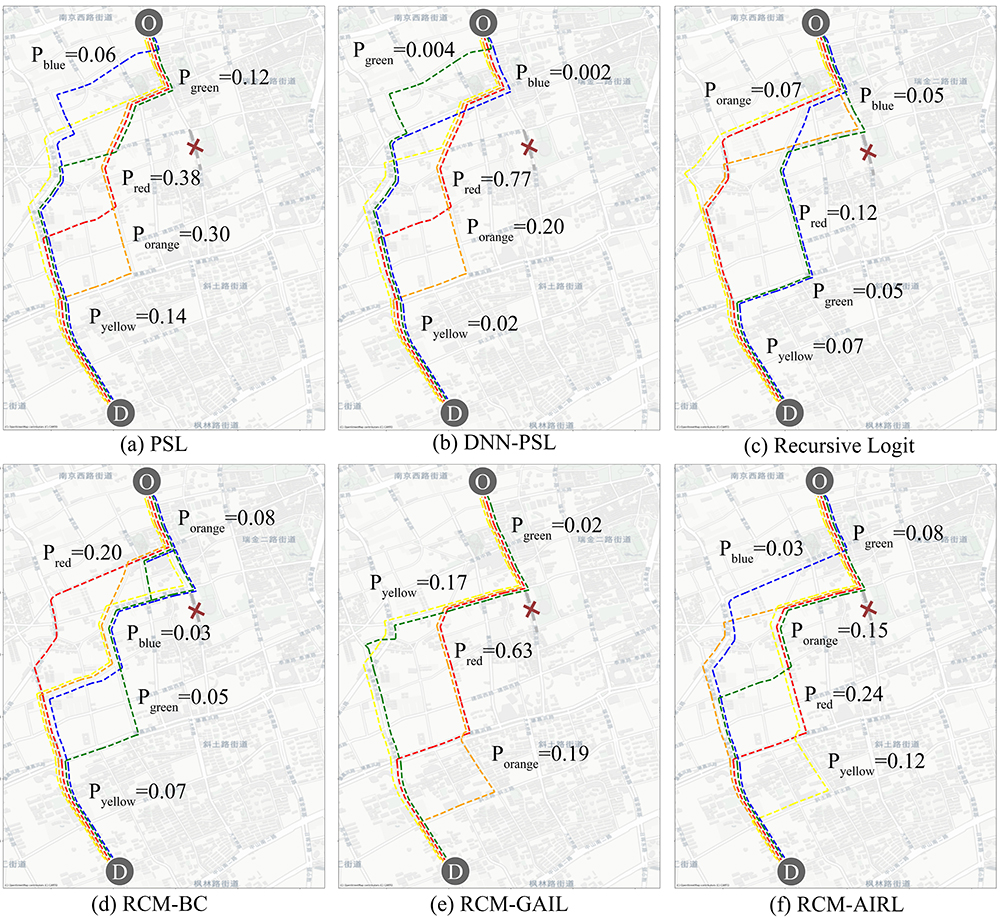}
  \caption{Prediction results of different models when a link is closed (the closed link is denoted as a grey line with a red cross)}\label{fig:counterfactual}
\end{figure}

One important use case of RCMs is for counterfactual analysis under hypothetical situations, e.g., adding or removing a link in the network. As the features used in this study are not network-specific, the proposed model should be applicable even when the network is updated, just as the existing RCMs. This would only require recomputing features based on the updated network, without the need to collect new data or retrain the model. In this section, we consider a scenario when a link is closed and predict how the route choice behavior would change using different models. Specifically, the input features for different models are reconstructed based on the updated network, and then we implement the models pre-trained based on the original network. The predicted route choice behavior for a selected OD pair is demonstrated in Figure~\ref{fig:counterfactual}. The chosen OD is the same as the OD used in Figure~\ref{fig:visualize}. By comparing the two figures, we can see how the route choice behavior would adapt based on the updated network, as predicted by the different models. Since the ground-truth data under this link closure scenario is not available, we cannot directly benchmark the model performance. Nevertheless, intuitively, the behavior of RCM-AIRL is as expected, and its applicability for counterfactual analysis should be comparable with existing methods.

\section{The effect of destination features}\label{appendix:c}
Figure~\ref{fig:shap} shows that destination-related context features are among the most important features in route choice modeling, with the shortest path distance to destination as the most important one, at least for the proposed RCM-AIRL. To further quantify the contribution of destination features, we implement a variant model of RCM-AIRL without these features. We conduct experiments using a training data size of 10000 trips based on 5-fold cross validation. The performance comparison with and without destination features is displayed in Table~\ref{table:des_ablate}. It is found that without the destination features, the performance of the variant model is quite poor. Further analysis reveals that the trajectories generated by the variant model can easily get stuck in a loop and never reach the desired destination. The likely reason is that the reward function learned in the variant model only depends on state-action pairs, ignoring the fact that a traveler's route choice preference is closely related to the desired destination. In a model-free method such as RCM-AIRL, these destination-related features appear to be crucial to encourage goal-oriented exploration and improve learning efficiency.

\begin{table}[ht!]
  \centering \footnotesize
  \caption{Route choice prediction performance of RCM-AIRL with and without destination features}
    \begin{tabular}{cccccc}
    %\addlinespace
    \toprule
    With/without destination features & ED & BLEU & JSD & LP\\
    \midrule
    Without destination features & 0.829 & 0.262 & 0.593 & -9.018\\
    With destination features & 0.228 & 0.717 & 0.351 & -4.301\\
    \bottomrule
    \end{tabular}%
  \label{table:des_ablate}%
\end{table}%

\section{Interpretability of baseline models}\label{appendix:d}
\begin{figure}[!ht]
  \centering
  \includegraphics[width=0.8\textwidth]{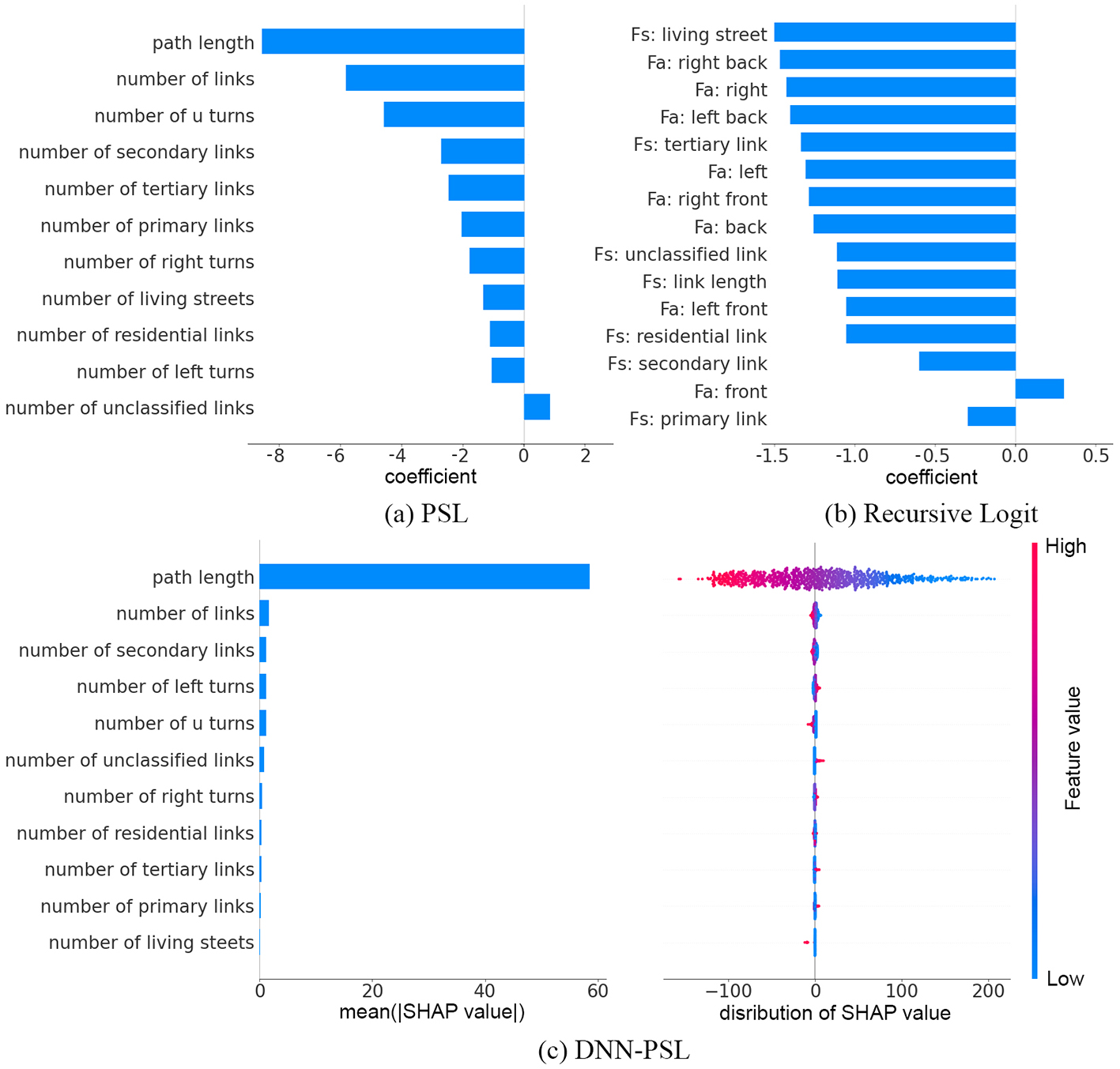}
  \caption{Interpretation of baseline models}\label{fig:interpret_baseline}
\end{figure}

In this section, we conduct further interpretation analysis to examine the effect of different features for different baseline models. Since PSL and recursive logit assume linear relationships between explanatory features and routing preferences, we report their estimated coefficients. For deep learning-based models such as DNN-PSL, we use SHAP to understand how the input features influence their estimation results. The results for DNN-BC and RCM-GAIL are not shown, because they do not have a specific model structure to estimate reward functions. All the interpretation results are displayed in Figure~\ref{fig:interpret_baseline}. Note that due to the different properties of different models, their input features can be different: PSL and DNN-PSL use path-level features as input, while recursive logit requires link-additive features related to states as input. Compared with the traditional baselines, our introduced approaches are more flexible to consider various features. Path-based models identify path length (or the distance to the destination) to be the most important features and travelers prefer routes with shorter distance, which is consistent with the finding of RCM-AIRL. Recursive logit identifies living street as the most important feature. A possible explanation is that living streets often limit the speed of cars and thus the attribute of living street can potentially reflect the travel cost of a trip.

%\section*{References}
\bibliographystyle{model5-names2}\biboptions{authoryear}
\bibliography{ref}






% \end{thebibliography}

\end{document}